\documentclass{article} 
\usepackage{baai,times}

\usepackage{hyperref}
\usepackage{url}
\usepackage{graphicx}
\usepackage{subcaption}
\usepackage{amssymb}
\usepackage{marvosym}
\usepackage{array}
\usepackage{enumitem}
\usepackage{booktabs}
\usepackage{tabularx}
\usepackage[most]{tcolorbox}
\usepackage{placeins}
\usepackage{tikz}
\usetikzlibrary{positioning, shapes.geometric, arrows.meta, fit, calc, backgrounds}

\title{\texorpdfstring{%
  \makebox[\textwidth][c]{\includegraphics[height=3em]{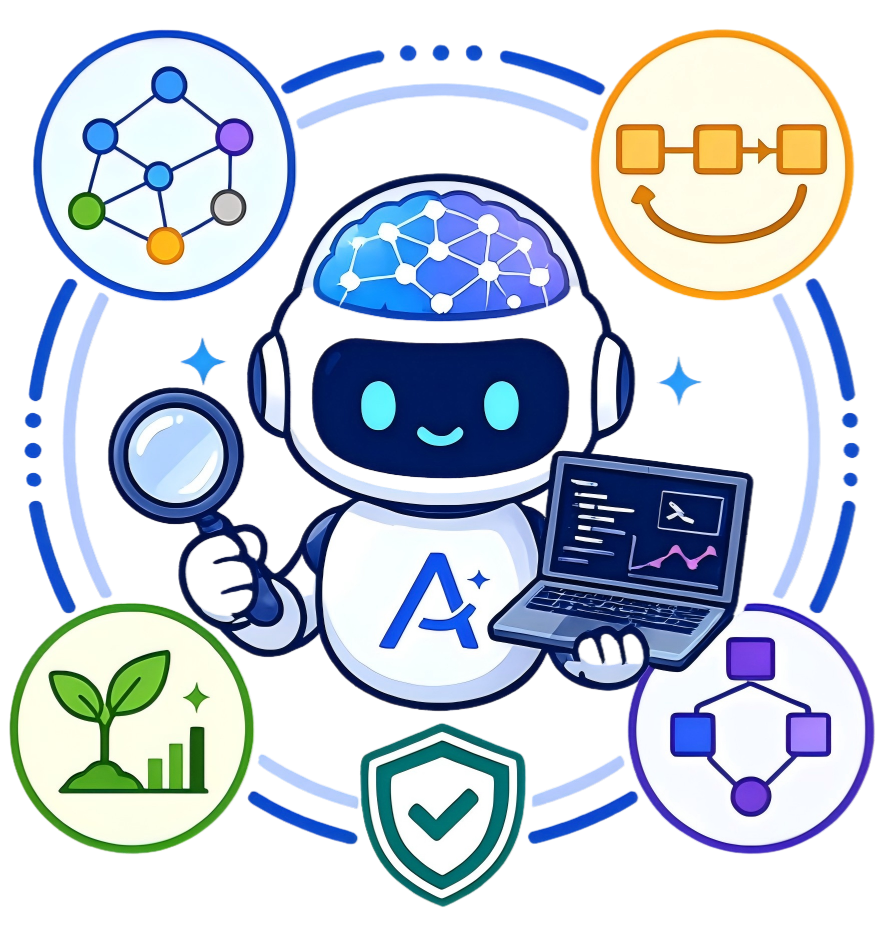}}\\[0.5em]
  AutoSci: A Memory-Centric Agentic System for the Full Scientific Research Lifecycle%
  }{%
  AutoSci: A Memory-Centric Agentic System for the Full Scientific Research Lifecycle%
  }}

\author{\normalfont
\small\bfseries Weitong Qian\textsuperscript{*,\textdagger},
Beicheng Xu\textsuperscript{*,\textdagger},
Zhongao Xie\textsuperscript{*,\textdagger},
Bowen Fan\textsuperscript{\textdagger},
Guozheng Tang\textsuperscript{\textdagger},
Jiale Chen\textsuperscript{\textdagger},\\
\small\bfseries
Xinzhe Wu\textsuperscript{\textdagger},
Mingtian Yang\textsuperscript{\textdagger},
Chenyang Di\textsuperscript{\textdagger},
Jiajun Li\textsuperscript{\textdagger},
Lingching Tung\textsuperscript{\textdagger},
Peichao Lai\textsuperscript{\textdagger},\\
\small\bfseries
Yifei Xia\textsuperscript{\textdagger},
Ziyi Guo\textsuperscript{\textdagger},
Yanwei Xu\textsuperscript{\textdagger},
Yanzhao Qin\textsuperscript{\textdagger},
Shaoduo Gan\textsuperscript{\textdagger},
Xupeng Miao\textsuperscript{\textdagger},
Bin Cui\textsuperscript{\textdagger,\Letter}\\
\normalfont\small
\textsuperscript{\textdagger}Peking University, Beijing, China\\
\normalfont\small
\textsuperscript{*}Equal contribution. \textsuperscript{\Letter}Corresponding author.\\
\normalfont\small
\textsuperscript{*}\{weitong.qian, beichengxu, zaxie25\}@stu.pku.edu.cn, \textsuperscript{\Letter}bin.cui@pku.edu.cn
}

\iclrfinalcopy 
\begin{document}


\maketitle

\begin{abstract}
Scientific research has traditionally been human-intensive, requiring researchers to coordinate literature, ideas, experiments, manuscripts, and review responses across long project cycles. 
The rise of LLM-based scientific agents creates an opportunity to automate this process. 
Such a system must support the full research lifecycle, maintain structured persistent memory across projects, and improve its own research procedures over time.
However, existing systems either partially satisfy or fail to satisfy these requirements, leaving a gap for a unified automated scientific research system.
As a result, we present AutoSci, a memory-centric agentic system for the full scientific research lifecycle. 
AutoSci is organized around four modules. SciMem provides schema-governed research memory, 
 separating Long-Term Knowledge Memory for reusable scientific knowledge from Active Research Memory for project-level artifacts such as ideas, experiments, manuscripts, and reviews. 
SciFlow executes a five-stage lifecycle from literature understanding to rebuttal through a harness that controls state, context, verification, feedback, and orchestration. 
SciDAG augments difficult skills with DAG-shaped multi-agent operators and reusable stage-specific templates. 
SciEvolve converts feedback signals from users, experiments, reviews, and external environments into versioned updates to SciMem organization, SciFlow skills, and SciDAG templates. 
Together, these modules make AutoSci a persistent research environment that can execute, remember, and evolve across research projects.
The code repository is available at \url{https://github.com/skyllwt/AutoSci}.
\end{abstract}

\section{Introduction}

Scientific research has long been a heavily human-driven process: 
researchers must manually track literature, formulate hypotheses, implement methods, run experiments, analyze evidence, write papers, and respond to reviews. 
This process is labor-intensive, especially when projects require broad literature coverage, experimentation, and careful coordination across many intermediate artifacts. 
The rise of large language models and multi-agent systems has begun to change this picture. 
When coupled with tools, code execution, external scientific resources, and coordinated workflows, these systems can automate the research lifecycle and support systematic exploration, validation, monitoring, and manuscript
  production~\citep{huang2025popper,chai2025scimaster,zhang2026autodeepresearcher}.


One line focuses on particular scientific capabilities rather than a whole paper-production pipeline. The AI co-scientist~\citep{gottweis2025coscientist} targets scientist-in-the-loop hypothesis generation and biomedical validation; 
  POPPER~\citep{huang2025popper} studies automated falsification of free-form hypotheses; 
  AutoSciLab~\citep{desai2024autoscilab} develops self-driving laboratory workflows; 
  bilevel LLM--simulation optimization connects LLM reasoning with scientific simulation~\citep{ma2024llmsimulation}; 
  SciMaster/X-Master~\citep{chai2025scimaster} uses code and tool-augmented reasoning for scientific problem solving; 
  and Deep Researcher Agent~\citep{zhang2026autodeepresearcher} emphasizes sustained experiment execution, monitoring, and reflection. 
These works demonstrate the value of LLM agents for individual scientific operations, but they do not by themselves define a complete research lifecycle.


More ambitious systems move beyond individual capabilities and target complete research workflows. 
The AI Scientist series~\citep{lu2024aiscientist,yamada2025aiscientistv2} and AI-Researcher~\citep{tang2025ai} automate idea generation, experiment execution, paper writing, review, and later template-free agentic
  search; while Agent Laboratory~\citep{schmidgall2025agentrxiv} supports research workflows starting from a user-provided idea. 
Related full-loop systems further model research--review--refinement, goal-oriented  cumulative findings, and memory-augmented discovery, as in CycleResearcher~\citep{weng2024cycleresearcher}, DeepScientist~\citep{weng2025deepscientist}, and EvoScientist~\citep{lyu2026evoscientist}. 
As these workflows become longer-running, research harness design becomes central, since scientific agents need durable state, tool contracts, review gates, and recoverable execution in addition to stronger models. 
Representative harness-oriented systems include ARIS~\citep{yang2026aris}, NORA~\citep{zhou2026nora}, and Deep Researcher Agent~\citep{zhang2026autodeepresearcher}, which add persistent state, monitoring, provenance or claim checks, review
  gates, domain guardrails, and recoverable long-running execution.
These systems move scientific agents toward end-to-end automation, but most
  remain organized around a single project or a paper-generation pipeline.

\begin{table*}[t]
\centering
\footnotesize
\newcommand{\full}{\raisebox{-0.15ex}{\large\textcolor{green!55!black}{\checkmark}}}
\newcommand{\partly}{\raisebox{-0.1ex}{\large\(\circ\)}}
\newcommand{\none}{--}
\newcommand{\syscite}[2]{{\renewcommand{\arraystretch}{1.0}\begin{tabular}[t]{@{}l@{}}#1\\[-1pt]\citep{#2}\end{tabular}}}
\setlength{\tabcolsep}{7pt}
\renewcommand{\arraystretch}{1.5}
\begin{tabular}{p{0.34\textwidth}cccc}
\hline
\textbf{System} &
\textbf{Harness} &
\begin{tabular}{c}\textbf{Structured}\\\textbf{Sci. Mem.}\end{tabular} &
\begin{tabular}{c}\textbf{Persistent}\\\textbf{Sci. Mem.}\end{tabular} &
\begin{tabular}{c}\textbf{System}\\\textbf{Evolution}\end{tabular} \\
\hline
\syscite{AI Scientist series}{lu2024aiscientist,yamada2025aiscientistv2} & \partly & \none & \none & \none \\
AI-Researcher~\citep{tang2025ai} & \partly & \none & \none & \none \\
\syscite{Agent Laboratory}{schmidgall2025agentrxiv} & \partly & \none & \none & \none \\
CycleResearcher~\citep{weng2024cycleresearcher} & \partly & \none & \none & \none \\
EvoScientist~\citep{lyu2026evoscientist} & \partly & \partly & \full & \partly \\
DeepScientist~\citep{weng2025deepscientist} & \full & \partly & \partly & \none \\
ARIS~\citep{yang2026aris} & \full & \partly & \partly & \none \\
NORA~\citep{zhou2026nora} & \full & \partly & \partly & \none \\
\syscite{Deep Researcher Agent}{zhang2026autodeepresearcher} & \full & \partly & \partly & \none \\
\textbf{AutoSci} & \textbf{\full} & \textbf{\full} & \textbf{\full} & \textbf{\full} \\
\hline
\end{tabular}
\caption{Feature-level comparison of representative full-loop scientific-agent systems and research harnesses. Symbols denote full support (\textcolor{green!55!black}{\checkmark}), partial or project-local support (\(\circ\)), and features that are not a primary focus (--). Here, persistent memory refers to research memory that survives across complete research or paper-generation pipelines and can be reused by later pipelines, rather than state retained only within a single run. Full system evolution further requires modifying the scientific-agent system itself, such as its skills, workflow protocols, prompts, or artifact contracts; accumulating reusable textual experience alone is counted as partial support.}
\label{tab:scientific-agent-features}
\end{table*}

From a system perspective, an automated scientific research system should satisfy four requirements:
1) \textbf{Full-lifecycle support}. Since scientific research spans literature understanding, idea generation, experimental validation, manuscript writing, and rebuttal, an automated system should provide stage-specific skills and artifact
  handoffs throughout the entire process. 
  Table~\ref{tab:scientific-agent-features} therefore focuses on representative full-loop scientific agents and compares how well they satisfy the remaining system requirements.
2) \textbf{Execution harness}. Long-running research cannot rely on unconstrained conversations alone; it requires persistent state, controlled context, verification gates, feedback routing, and recoverable orchestration. 
  However, some earlier systems mainly coordinate research with lighter runtime control.
3) \textbf{Structured and persistent memory}. Memory must be structured to make scientific information semantically interpretable, extensible, and organized by its dependencies rather than stored as undifferentiated text.
However, existing systems mostly store summaries, logs, strategies, or artifacts rather than organizing scientific information as typed objects with explicit dependencies. 
Moreover, most prior systems retain memory only within a single research project or paper-generation pipeline, rather than preserving reusable cross-project experience for future workflows.
4) \textbf{Self-evolution}. An automated research system should not only accumulate textual experience, but also use user feedback and experimental outcomes to iteratively improve its own skills and workflows. 
Although EvoScientist distills prior experience into reusable textual memories, existing systems generally do not revise the agent system itself, such as its skills and workflow protocols.
Overall, the comparison shows that existing systems remain fragmented: they improve different parts of automated science, but do not yet form a unified research system that can execute, remember, and evolve across projects.

\begin{figure*}[t!]
    \centering
    \includegraphics[width=\textwidth]{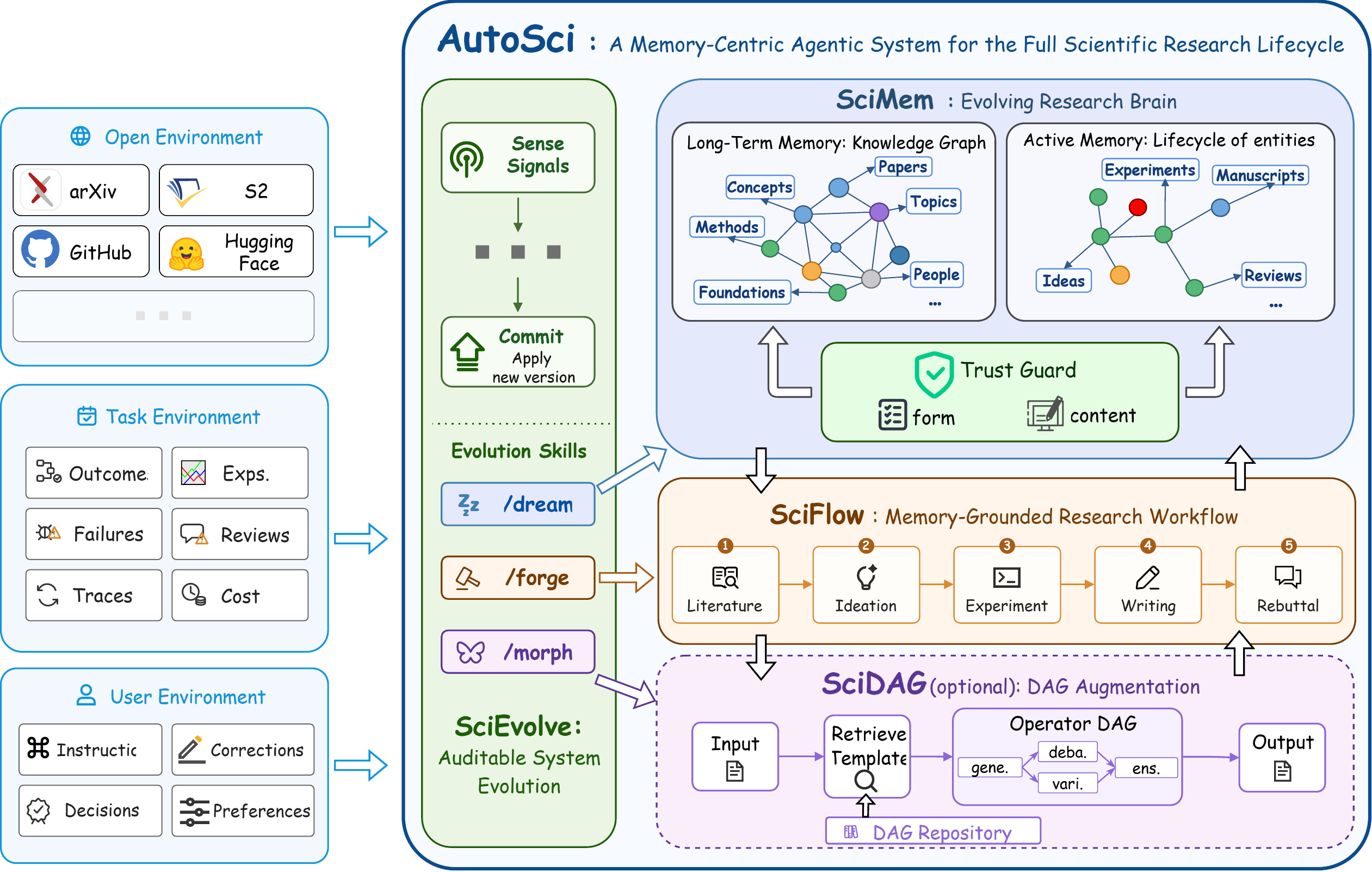}
    \caption{Overview of AutoSci.}
    \label{fig:autosci-overview}
\end{figure*}

To address this gap, we introduce \emph{AutoSci}, a memory-centric agentic system for the full scientific research lifecycle. 
AutoSci is built around four modules: 
  \emph{SciMem}, a structured persistent research memory that stores scientific knowledge, active project artifacts, and cross-project experience; 
  \emph{SciFlow}, a harness-based execution framework for the full scientific research lifecycle, coordinating literature, ideation, experimentation, writing, and rebuttal; 
  \emph{SciDAG}, a DAG-based multi-agent augmentation mechanism for stages that require broader search, debate, verification, or refinement; 
  and \emph{SciEvolve}, a full-system evolution layer that converts user feedback, experimental outcomes, and review signals into versioned updates to memory, skills, and orchestration templates.
Figure~\ref{fig:autosci-overview} provides an overview of the AutoSci architecture.
Our main contributions are summarized as follows:
  \begin{itemize}[leftmargin=1.5em,itemsep=1pt,topsep=1pt]
      \item We formulate automated scientific research as a long-lifecycle system problem that requires execution, memory, and evolution across projects, rather than isolated task automation.
      \item We design a memory-centric architecture in which structured persistent scientific memory serves as the shared substrate for research workflows, multi-agent execution, and self-improvement.
      \item We implement an end-to-end AutoSci system that integrates lifecycle skills, harnessed execution, DAG-based multi-agent augmentation, and versioned self-evolution.
      \item We conduct two end-to-end case studies in GPU kernel optimization and biomedical drug discovery, where AutoSci generates reviewable paper-level artifacts that receive automated ICLR-review scores of 6.3/10 and 5.8/10, respectively.
  \end{itemize}

\section{System Overview}
\label{sec:method}

AutoSci targets long-lifecycle scientific research: a system should not only complete one research project, but also accumulate knowledge, experimental experience, submission feedback, and execution strategies across many projects. This goal requires the agent to behave less like a single-session assistant and more like a persistent research environment that can interact with users, external resources, and experimental systems over time.
We follow four design principles:
\begin{itemize}[leftmargin=2em,itemsep=1pt,topsep=2pt]
    \item \textbf{Environment interaction.} AutoSci should interact with the full research environment, including user instructions, literature and codebases, and the experimental runtime.
    \item \textbf{Structured persistent memory.} AutoSci should maintain structured long-term scientific memory, so that papers, concepts, methods, experiments, reviews, and their relations can persist and be reused across complete research projects.
    \item \textbf{Harnessed execution.} AutoSci should run through an explicit harness to make the research lifecycle interruptible, reviewable, and reusable across sessions..
    \item \textbf{Full-system evolution.} AutoSci should go beyond accumulating reusable experience by turning recurring feedback into controlled updates to its own skills, protocols, and prompts.
\end{itemize}

\begin{table}[t!]
\centering
\caption{AutoSci system overview (v1.0.0, May 2026).}
\label{tab:autosci_overview}
\begin{tabularx}{\textwidth}{l X}
\toprule
\textbf{Component} & \textbf{Scope} \\
\midrule
System modules & 4 modules: SciMem, SciFlow, SciDAG, and SciEvolve \\
Persistent memory & 2 regions (Long-Term Knowledge + Active Research); 10 typed entities; 20+ typed relations \\
Research workflows & /research (Literature $\rightarrow$ Ideation $\rightarrow$ Experiment $\rightarrow$ Writing $\rightarrow$ Rebuttal) \\
Research Skills & 30+ skill, spanning the five lifecycle stages and self-evolution \\
Multi-agent operators & 9 reusable operators (generate, variation, debate, refine, review, etc.) \\
Self-evolution & 3 evolution skills \\
\bottomrule
\end{tabularx}
\end{table}

These principles lead to four modules. \emph{SciMem} provides schema-governed research memory; \emph{SciFlow} executes the full scientific lifecycle over that memory; \emph{SciDAG} optionally augments difficult stages with DAG-shaped multi-agent operators; and \emph{SciEvolve} converts traces and feedback into versioned system updates. Together, the modules form a closed loop in which research artifacts are produced, checked, stored, reused, and eventually used to improve the system itself. Table~\ref{tab:autosci_overview} summarizes the implemented system scope before we describe each module in detail.

\section{SciMem: Schema-Governed Research Memory}

SciMem is designed for long-lifecycle scientific memory: memory should not disappear after a single experiment, paper, or research project, but should remain reusable across future projects. 
To support this goal, SciMem separates memory into two regions with different responsibilities. \emph{Long-Term Knowledge Memory} preserves consolidated scientific knowledge that should accumulate across projects, while \emph{Active Research Memory} tracks the fast-changing state of an ongoing research paper or experimental report. 
Below, we first introduce the two memory regions and then describe how SciMem grows and flows across them over time.

\subsection{Long-Term Knowledge Memory}


Long-Term Knowledge Memory is to preserve the scientific knowledge that AutoSci has accumulated from external sources and prior research cycles, so that later projects can reuse it without reconstructing the same context from scratch. The region is populated by literature ingestion skills from sources such as arXiv, Semantic Scholar, GitHub, and user-provided documents, and can be refined by consolidated experience from completed projects.
The region is organized by a typed entity schema rather than by flat documents or vector chunks. 
Table~\ref{tab:ltkm-entities} summarizes the long-term entity types. \emph{In implementation}, each entity is stored as a \texttt{.md} page.

\begin{figure*}[t]
    \centering
    \begin{subfigure}[t]{0.42\textwidth}
        \centering
        \includegraphics[width=\linewidth]{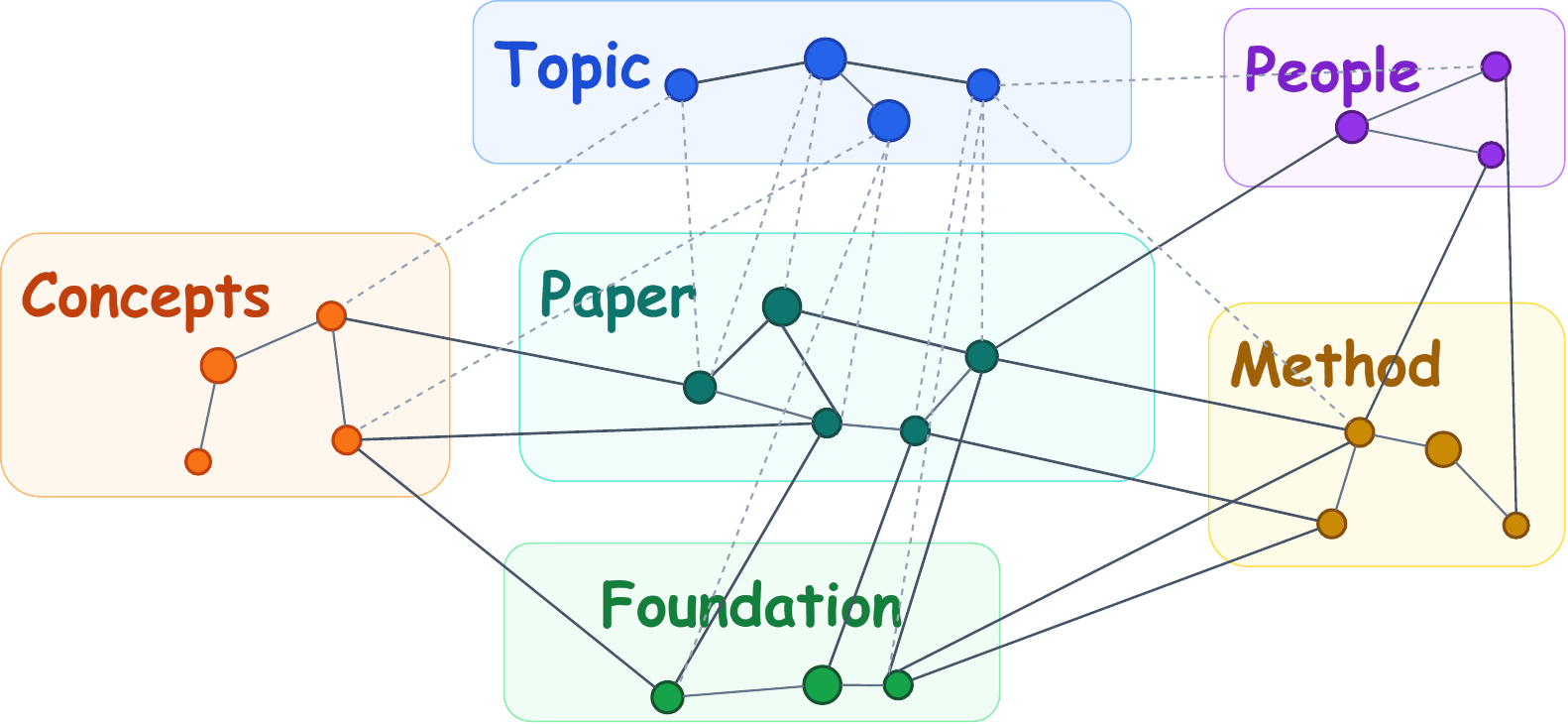}
        \caption{Long-Term Knowledge Memory.}
        \label{fig:scimem-ltm}
    \end{subfigure}
    \hfill
    \begin{subfigure}[t]{0.55\textwidth}
        \centering
        \includegraphics[width=\linewidth]{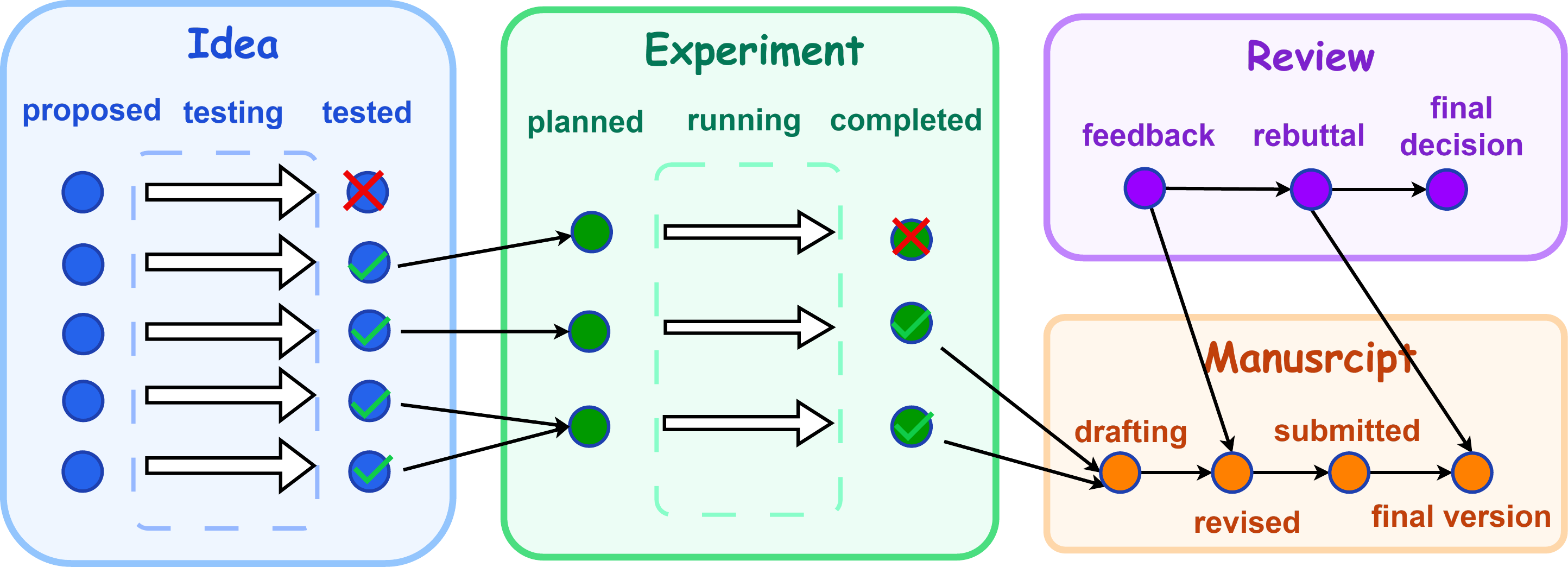}
        \caption{Active Research Memory.}
        \label{fig:scimem-arm}
    \end{subfigure}
    \caption{Two memory regions in SciMem.}
    \label{fig:scimem-regions}
\end{figure*}

Beyond defining entity types, the long-term schema also governs how these entities are connected. 
For example, \texttt{Topic} entities provide the coarsest organizing layer: \texttt{Paper}, \texttt{Foundation}, \texttt{Concept}, \texttt{Method}, and \texttt{People} entities can be placed within one or more topics. 
\texttt{Paper} entities act as evidence-bearing sources that introduce or critique \texttt{Concept} entities, apply or extend \texttt{Method} entities, and contribute to \texttt{Foundation} entities. 
\texttt{Foundation} entities provide stable background knowledge that grounds \texttt{Concept} and \texttt{Method} entities. 
These typed relations turn Long-Term Knowledge Memory from a set of structured pages into a traversable scientific knowledge graph. 
\emph{In implementation}, typed links are stored as schema-constrained bidirectional cross-references between entity pages, making the graph navigable and mechanically checkable.
Figure~\ref{fig:scimem-ltm} visualizes the entity schema and typed connections of Long-Term Knowledge Memory.

Long-Term Knowledge Memory has two defining properties: 
 1) \emph{semantic addressability}, which lets downstream skills retrieve typed scientific objects and relations directly; 
 and 2) \emph{incremental extensibility}, which lets new
  literature and validated findings be appended across research pipelines, making the memory a reusable scientific substrate rather than a project-local cache.

\begin{table}[t]
\centering
\footnotesize
\setlength{\tabcolsep}{4pt}
\renewcommand{\arraystretch}{1.12}
\begin{tabular}{p{0.16\linewidth}p{0.7\linewidth}}
\hline
\textbf{Entity} & \textbf{Typical Content} \\
\hline
\texttt{Topic} & Domain scope, key observations \\
\texttt{Paper} & Structured reading notes that capture the essence of the paper \\
\texttt{Foundation} & Consolidated background knowledge used as a stable basis \\
\texttt{Concept} & Reusable descriptions of scientific notions or terminology \\
\texttt{Method} & Detailed implementation and functional role of a reusable technical approach \\
\texttt{People} & Research profiles of researchers \\
\hline
\end{tabular}
\caption{Entity types in Long-Term Knowledge Memory.}
\label{tab:ltkm-entities}
\end{table}

\subsection{Active Research Memory}

Active Research Memory is the project-level workspace for producing a research paper or experimental report from start to finish. It records the key active artifacts of the current project, including \texttt{Idea}, \texttt{Experiment},
  \texttt{Manuscript}, and \texttt{Review} entities, as AutoSci advances the project.
Each active entity carries an explicit lifecycle state. \texttt{Idea} entities move from \texttt{proposed} to \texttt{testing}, then to \texttt{tested}, \texttt{validated}, or \texttt{failed}; \texttt{Experiment} entities move from \texttt{planned} and \texttt{running} to \texttt{completed} or \texttt{abandoned}; \texttt{Manuscript} entities move through \texttt{drafting}, \texttt{revised}, \texttt{submitted}, and \texttt{final version}; and \texttt{Review} entities record received feedback, rebuttal drafting, revision, and final decision. These lifecycle states make Active Research Memory a structured progress map rather than a loose project folder.
\emph{In implementation}, each active entity is stored as a \texttt{.md} page with a schema-defined lifecycle state.
Figure~\ref{fig:scimem-arm} visualizes the active entity schema and lifecycle structure of Active Research Memory.


The significance of recording Active Research Memory is that AutoSci can recover and audit the state of a research project without relying on chat history. 
At any moment, the system can identify which \texttt{Idea} entities are still viable, which \texttt{Experiment} entities produced evidence, and which \texttt{Review} concerns remain unresolved. 
Terminal active artifacts also become the bridge back to Long-Term Knowledge Memory as reusable knowledge for future projects.

\begin{figure*}[t]
    \centering
    \includegraphics[width=0.95\textwidth]{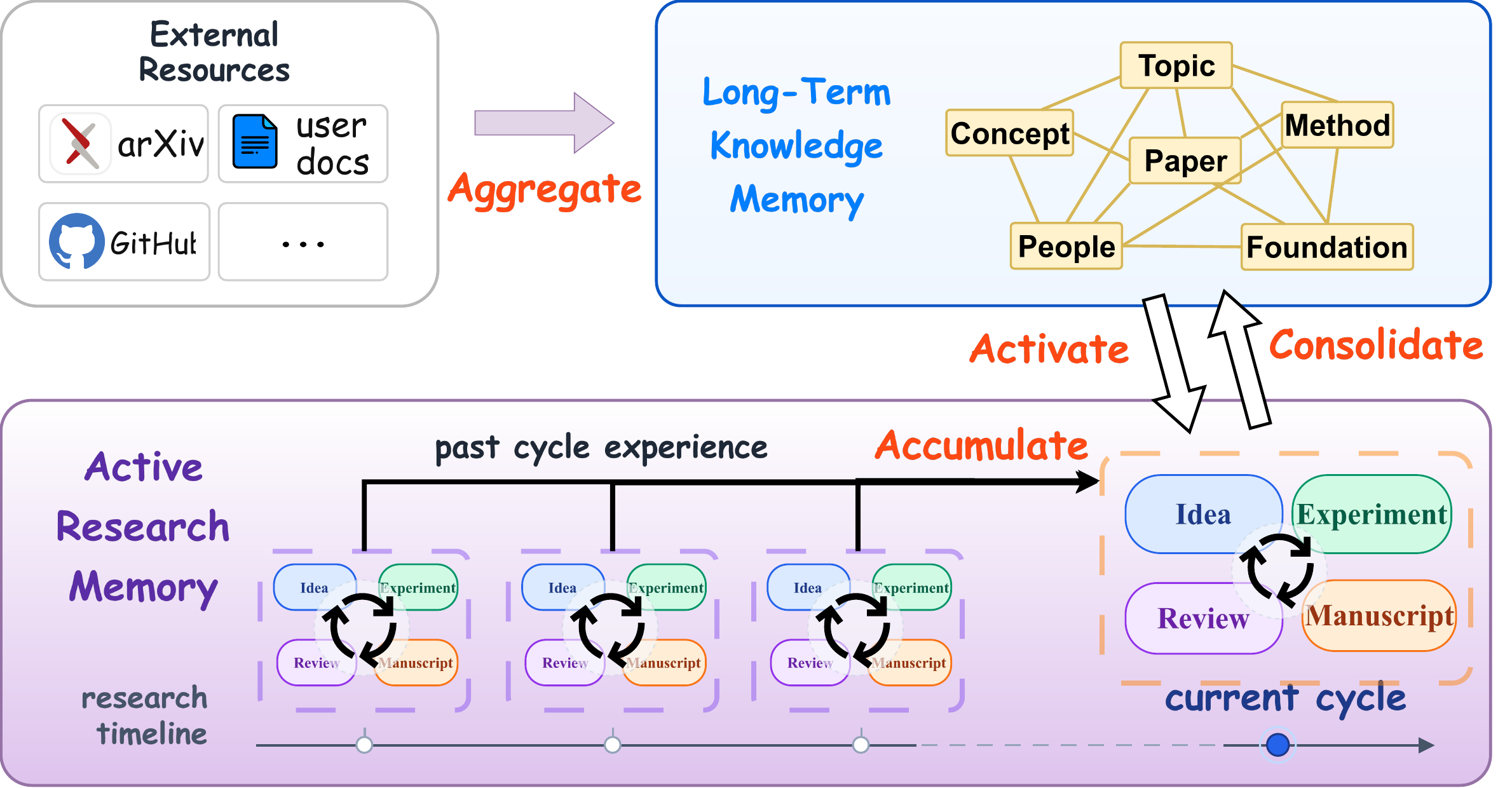}
    \caption{Memory growth and flow in SciMem.}
    \label{fig:scimem-flow}
\end{figure*}

\subsection{Memory Growth and Flow}

The previous sections define what SciMem stores. We next describe how the memory grows. SciMem expands through three complementary flow paths: aggregation within Long-Term Knowledge Memory, bidirectional flow between Long-Term Knowledge Memory and Active Research Memory, and temporal accumulation of cross-cycle experience, as illustrated in Figure~\ref{fig:scimem-flow}.

\textbf{Long-term aggregation.} Long-Term Knowledge Memory grows through internal aggregation. Newly ingested \texttt{Paper} entities do not remain isolated reading notes. Their key observations can update the domain-level understanding of a \texttt{Topic}; recurring definitions or mechanisms can refine a \texttt{Concept}; implementation details and empirical behavior can enrich a \texttt{Method}; and repeatedly supported background knowledge can strengthen a \texttt{Foundation}. This flow turns low-level source material into higher-level scientific memory, allowing long-term entities to become progressively more informative as more literature and project experience accumulate.

\textbf{Cross-region flow.} SciMem supports cross-region activation and consolidation. 
\textbf{1) Long-Term \(\rightarrow\) Active.} 
 During activation, an active entity draws on related long-term memory: a \texttt{Idea} entity is grounded in \texttt{Topic} entities, prior evidence, \texttt{Concept} entities, and \texttt{Method} entities; an \texttt{Experiment} entity activates the \texttt{Method} entities and assumptions it relies on; 
 and a \texttt{Manuscript} or \texttt{Review} entity activates the evidence it relies on. 
\textbf{2) Active \(\rightarrow\) Long-Term.} During consolidation, terminal active artifacts write back reusable scientific traces. \texttt{validated} \texttt{Idea} entities, completed experimental findings, failed attempts, and unresolved limitations can update the corresponding long-term entries, especially
  the relevant \texttt{Topic} pages.

\textbf{Cross-cycle accumulation.} SciMem accumulates methodological memory across cycles. Reviewer concerns and rebuttal outcomes are retained as cross-cycle notes that can be consulted when later projects enter writing or rebuttal stages. 
As a result, SciMem grows not only in what AutoSci knows about a topic, but also in how AutoSci conducts future research, experiments, writing, and rebuttal.

\textbf{Trust-guarded writes.}
  All SciMem writes pass through Trust Guard before entering the usable graph, since memory errors can propagate to future projects. 
  Trust Guard checks both \emph{form validity} (schema fields, lifecycle
  states, link types, and bidirectional links) and \emph{content validity} (evidence support and consistency with existing memory). 
  Form checks are handled by deterministic linting, while content checks are handled by an independent reviewer agent. Each write is assigned \textsc{Pass}, \textsc{Warn}, or \textsc{Block}; blocked artifacts are quarantined until resolved.
\section{SciFlow: Memory-Grounded Research Lifecycle}

\begin{figure*}[t]
    \centering
    \includegraphics[width=\textwidth]{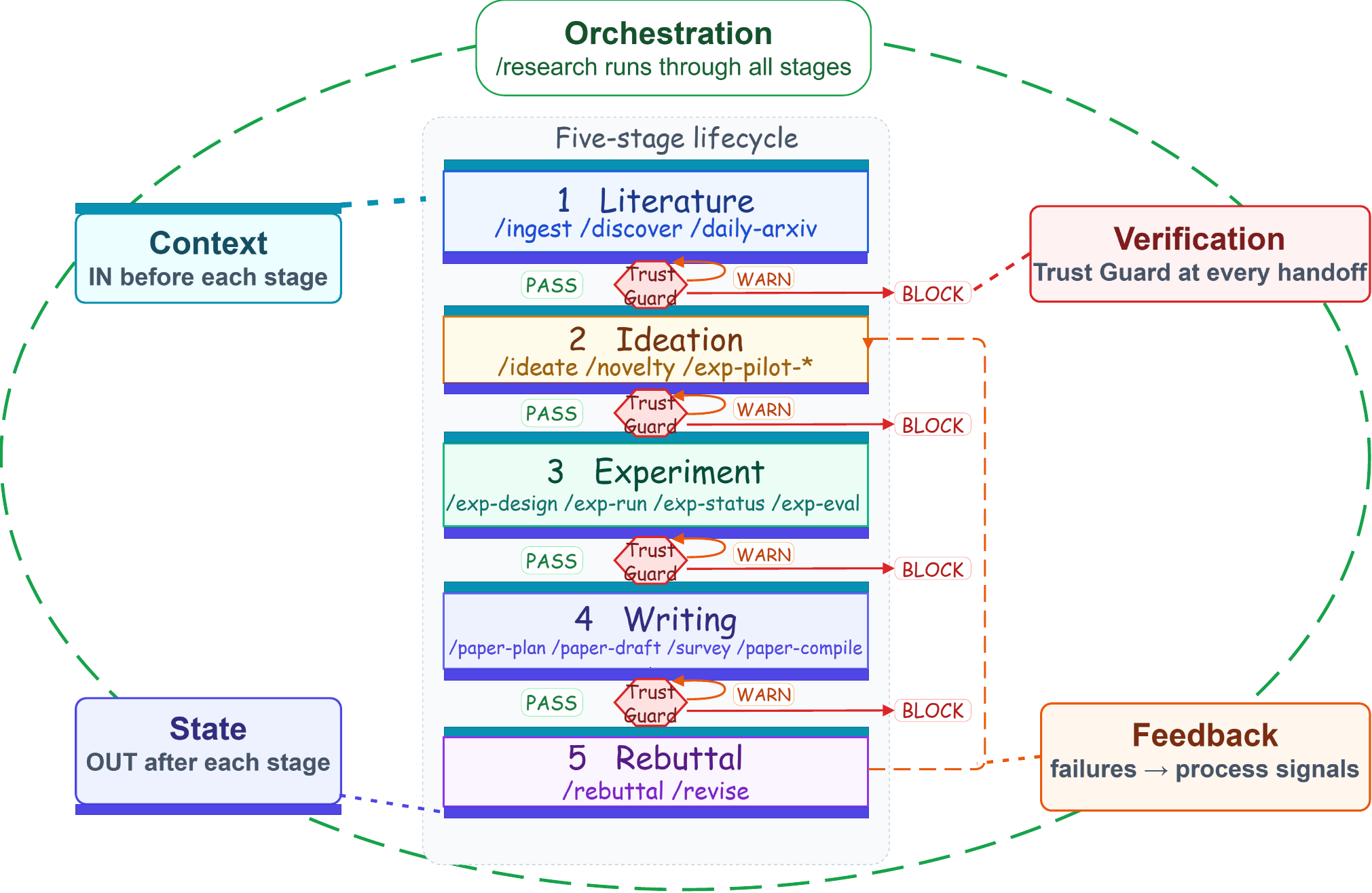}
    \caption{SciFlow research lifecycle and harness organization.}
    \label{fig:sciflow-harness}
\end{figure*}

SciFlow is the lifecycle executor that runs AutoSci over a complete research project. 
Its goal is to make long-horizon research executable, resumable, and memory-grounded rather than a sequence of free-form agent conversations. 
To this end, SciFlow decomposes a project into five stages, \emph{Literature}, \emph{Ideation}, \emph{Experiment}, \emph{Writing}, and \emph{Rebuttal}. 
Each stage is implemented as a harness-based skill contract.
Figure~\ref{fig:sciflow-harness} illustrates how the five-stage skills and harness guarantees are organized.

\subsection{Five-Stage Research Lifecycle}

SciFlow follows the natural lifecycle of a scientific project: 
it first builds the knowledge base, then proposes candidate directions, turns selected ideas into experimental evidence, writes the evidence into a paper, and finally handles reviewer feedback after submission. 
\emph{In implementation}, SciFlow is supported by more than 30 research skills spanning the five lifecycle stages.

\textbf{Memory-grounded execution.} 
The research lifecycle is memory-grounded because each stage is coupled to SciMem through explicit read and write operations. 
\emph{Literature} writes external knowledge into long-term memory. 
\emph{Ideation} reads long-term memory and writes \texttt{Idea} entities. 
\emph{Experiment} reads selected ideas, then writes evidence-bearing \texttt{Experiment} entities. 
\emph{Writing} reads provenance and evidence chains to produce \texttt{Manuscript} artifacts. 
\emph{Rebuttal} reads the submitted manuscript, review records, and prior rebuttal lessons, then writes new \texttt{Review} records. 
This read-write loop makes SciFlow memory-grounded: stages communicate through SciMem rather than transient conversation, and the memory used by later stages is already enriched by earlier stages.

\subsection{Harness Guarantees}

The five-stage lifecycle describes what scientific work is performed, while the SciFlow harness controls how this work is executed across stages. 
The harness is the cross-stage control layer around skills to make the lifecycle interruptible, reviewable, and reusable across sessions.

  \begin{itemize}[leftmargin=1.5em,itemsep=1pt,topsep=2pt]
      \item \textbf{State.} SciFlow records stage outputs, lifecycle states, links, and pipeline-level progress outside the transient LLM context, making projects resumable from a specified stage.
      \item \textbf{Context.} Before each skill runs, SciFlow equips it with a tailored SciMem view, providing the evidence, prior failures or lessons needed for that skill without exposing the full memory graph.
      \item \textbf{Verification.} Trust Guard checks memory writes and high-stakes handoffs through schema/link validation and evidence-oriented review before downstream stages consume them.
      \item \textbf{Feedback.} Failures and critiques are treated as process signals: insufficient evidence can trigger \texttt{/refine} or self-evolution.
      \item \textbf{Orchestration.} The \texttt{/research} loop invokes stage skills, records progress, handles stopping points, and supports long-running experiments through non-blocking execution and monitoring.
  \end{itemize}
\section{SciDAG: DAG-Based Multi-Agent Augmentation}

SciDAG is an optional augmentation for SciFlow. 
A selected skill can call a directed acyclic graph of reusable multi-agent operators as a tool to strengthen its execution. 
Given a stage task \(z\), the SciMem-compiled context, and the artifact schema expected by SciFlow, SciDAG executes an operator graph \(G=(V,E)\) and returns the final result to the same artifact
  contract, so downstream SciFlow stages remain unchanged.

\textbf{Adaptive operator graph.} SciDAG represents each tool call as an operator graph. 
Each node \(v_i \in V\) instantiates an operator \(o_i \in \mathcal{O}\) with a specialized sub-agent and produces an intermediate output from upstream node outputs.
Directed edges specify information flow, while conditional edges call a router over the current execution state to decide whether to continue, retry, branch, prune, or stop. 
Thus, SciDAG is not a fixed multi-agent chain: it adapts execution according to intermediate quality, cost, and convergence signals.

\textbf{Evolving templates.} To make such graphs reusable, SciDAG stores common operator graphs as stage-aware templates. 
 For example, Ideation templates emphasize diverse generation and debate, experimentation templates emphasize reliability checks, and writing templates emphasize evidence fidelity and refinement. 
 The template repository stores reusable graphs together with lightweight metadata and past execution experience. 
 For a new skill call, SciDAG retrieves a suitable template, executes it, and writes the resulting trace and feedback back to the repository.
Appendix~\ref{app:scidag-operators} lists the operator library and shows stage-specific templates for ideation, experimentation, and writing.

\section{SciEvolve: Full-System Evolution}

\begin{figure*}[t]
    \centering
    \includegraphics[width=\textwidth]{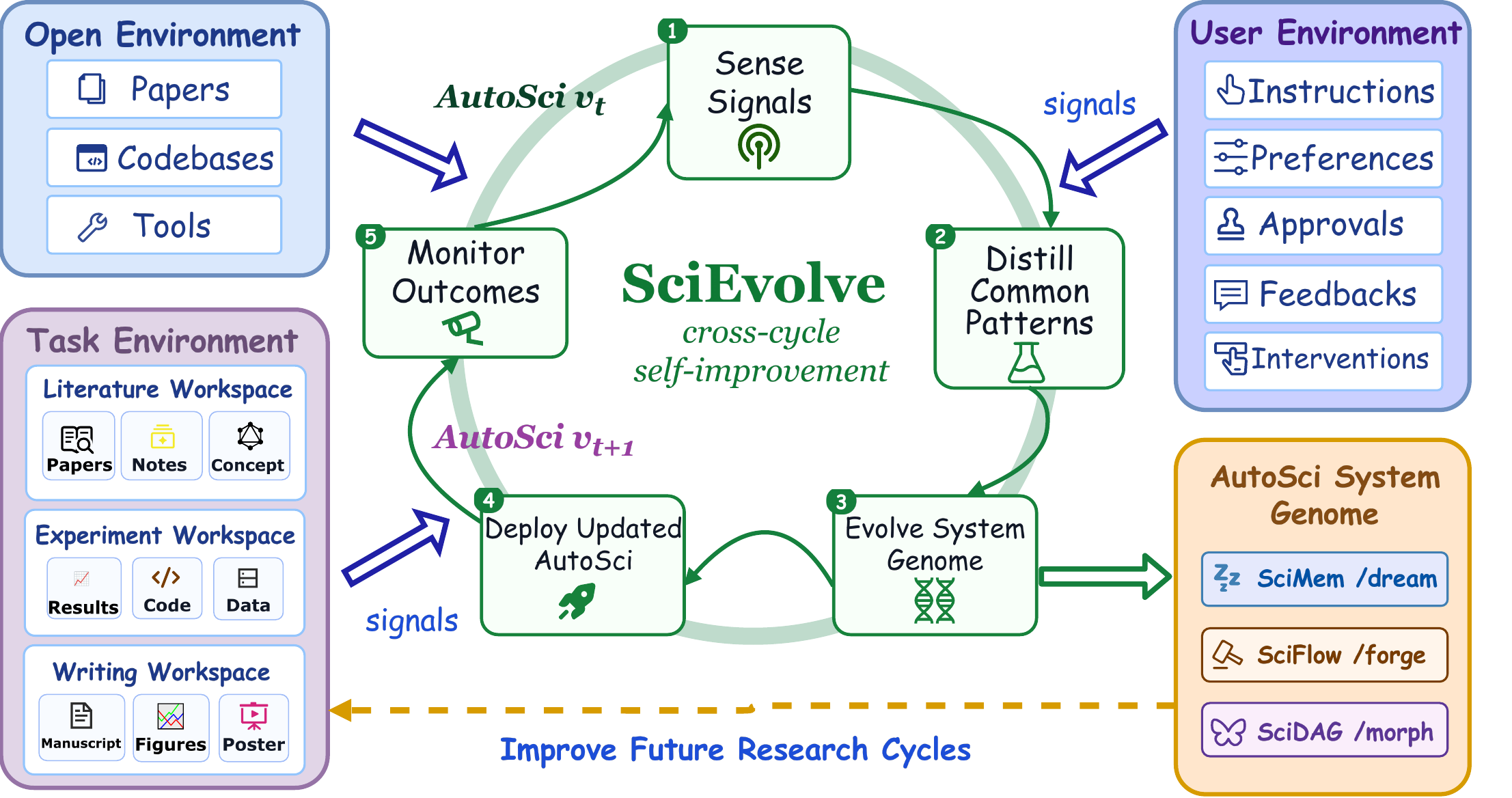}
    \caption{SciEvolve self-evolution loop.}
    \label{fig:scievolve}
\end{figure*}

SciEvolve implements the second part of AutoSci's self-improvement mechanism. 
Beyond the reusable textual experience accumulated through SciMem, SciEvolve turns feedback signals from research practice into auditable updates to SciMem organization, SciFlow skills, and SciDAG orchestration templates.
\emph{In implementation}, the three evolution paths are exposed as separate skills: \texttt{/dream} for SciMem evolution, \texttt{/forge} for SciFlow evolution, and \texttt{/morph} for SciDAG evolution.
Figure~\ref{fig:scievolve} illustrates this signal-to-update loop.

\textbf{Evolution signals.}
SciEvolve collects signals from three environments.
  The user environment provides instructions, corrections, and research preferences.
  The task environment provides stage outcomes, experimental evidence, and failure reasons.
  The open environment provides new papers, codebases, and venue expectations.
These signals are first stored in a signal repository, where SciEvolve detects recurring patterns and uses them to trigger updates to the relevant system module.
  
\textbf{SciMem evolution.}
Memory evolution maintains the usefulness of SciMem as it grows. 
\texttt{/dream} periodically reviews recent traces and related memory neighborhoods. 
It can down-weight or archive stale entries, compress redundant material, consolidate related entities, and propose new associations across \texttt{Concept}, \texttt{Method}, \texttt{Paper}, \texttt{Idea}, and \texttt{Experiment} entities.

\textbf{SciFlow evolution.}
Skill evolution treats SciFlow skills as versioned research protocols. 
A skill is not only a prompt, but a structured procedure that specifies inputs, required SciMem context, execution steps, checks, output artifacts, and handoff rules. 
After a research episode, SciEvolve analyzes repeated failure modes, user corrections, review warnings, unsupported claims, high-cost stages, and successful ad hoc repairs. 
When the evidence is stable enough, it proposes patches such as strengthening claim-evidence checks in writing skills, revising handoff requirements, or promoting a successful repair strategy into a reusable skill step. 

\textbf{SciDAG evolution.}
\texttt{/morph} uses SciDAG traces to improve multi-agent templates across executions.
  When an operator repeatedly underperforms, SciEvolve can revise its prompt, role, or tool configuration.
  When a graph shows stable failure or success patterns, SciEvolve can prune weak branches, add verification nodes, or specialize the template for a stage and problem type.
  Thus, SciEvolve improves SciDAG across executions.

\section{Case Studies and Evaluation}

\subsection{Experimental Setup}

We evaluate AutoSci through two end-to-end research case studies that span different scientific domains: GPU kernel optimization and biomedical drug discovery.
  The goal is not to test an isolated skill, but to examine whether AutoSci can run a complete research cycle, including literature organization, idea generation, novelty checking, feasibility analysis, experiment design, execution, result
  interpretation, and paper-oriented artifact production.
Both case studies use the same AutoSci system, including SciMem for persistent research memory, SciFlow for lifecycle execution, SciDAG for multi-agent augmentation when needed, and SciEvolve for recording reusable feedback signals.

\textbf{Case Study 1: GPU Kernel Optimization.}
  AutoSci explores iterative GPU operator optimization with Claude Code guided by performance feedback.
  The experiment is executed on a 4$\times$ NVIDIA A40 environment with Triton 3.2.0 and PyTorch 2.6.0+cu124 in the TritonBench workspace.
  
\textbf{Case Study 2: Biomedical Drug Discovery.}
  AutoSci explores structure-aware post-translational modification (PTM) modeling for degrader target nomination.
  The real executed blocks run on a single NVIDIA RTX 4060 using public DeepTernary v1.0.0 and PROTAC-STAN inference repositories, with additional Boltz-2-conditioned cross-checks on selected protein-of-interest cases.
  
For each case study, the user provides an initial research direction and a small set of relevant seed papers.
  We instantiate the AutoSci agents with Claude Code powered by Opus 4.7.
  The user then invokes the \texttt{/research} workflow, which first ingests the seed papers into SciMem, uses \texttt{/discover} to retrieve and ingest additional related papers, and constructs a structured Long-Term Knowledge Memory over
  papers, topics, concepts, methods, foundations, and researchers.
  After this memory-building stage, \texttt{/research} proceeds through ideation, novelty and feasibility screening, experiment design, experiment execution, result analysis, and manuscript-oriented artifact generation.
Because these case studies are simulated research submissions rather than papers undergoing real external peer review, we do not evaluate the rebuttal stage.

\subsection{Structured Memory Construction}

\begin{figure*}[t]
    \centering
    \includegraphics[width=0.98\textwidth]{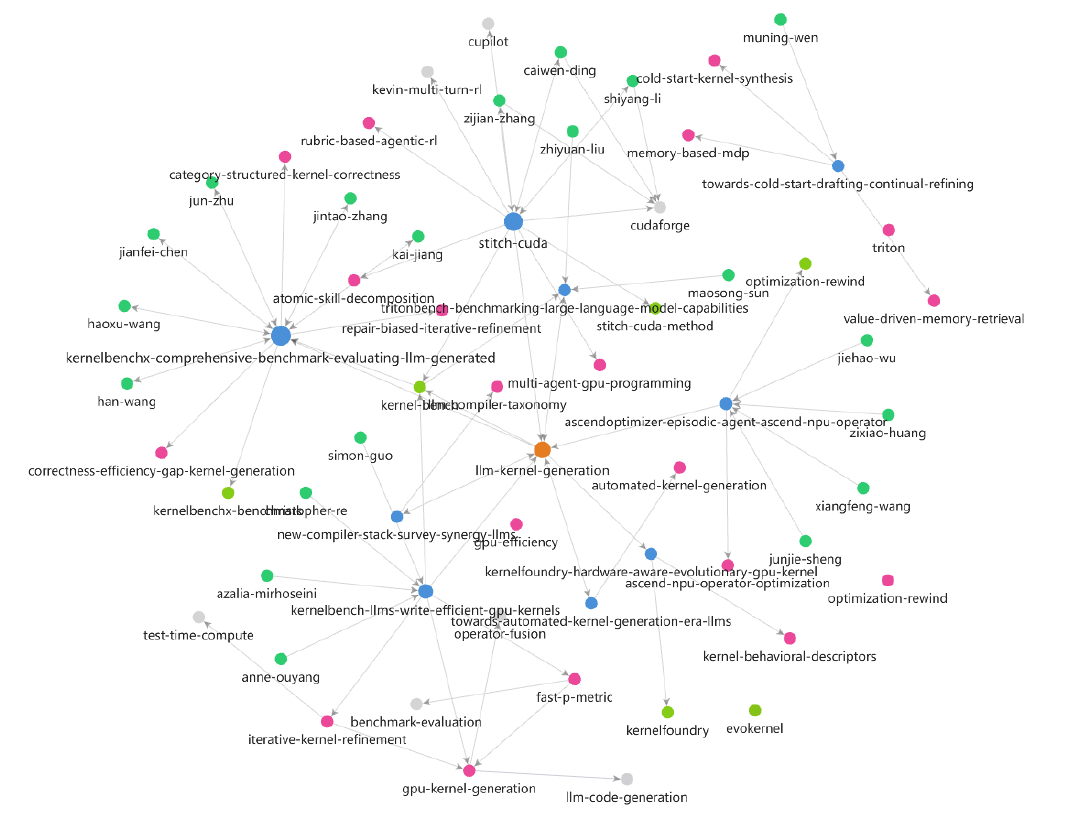}
    \caption{Example Long-Term Knowledge Memory built from the GPU kernel generation domain.}
    \label{fig:obsidian-ltkm}
\end{figure*}

We first examine whether AutoSci produces a structured and reusable Long-Term Knowledge Memory rather than a flat collection of notes.
Figure~\ref{fig:obsidian-ltkm} shows an example memory graph constructed in the GPU kernel optimization case study.
The graph contains typed entities such as topics, papers, concepts, methods, foundations, and researchers, with links recording how papers support concepts, how methods instantiate technical approaches, and how people connect to related research areas.
This structure allows later skills to retrieve scientific context by entity type and relation, instead of relying only on unstructured keyword search.
Appendix~\ref{app:ltkm-entity-examples} provides concrete entity-page examples from this memory.

\subsection{Idea Evolution and Selection}

We first analyze the idea pipeline in the GPU kernel optimization case study.
Given the user direction of iterative GPU operator optimization with Claude Code and performance feedback, AutoSci generates five candidate paths, performs novelty checking, refines the surviving ideas, evaluates feasibility under the project hardware budget, and selects one idea for full experimentation.
Figure~\ref{fig:kernel-idea-pipeline} summarizes this process.
Specifically, \texttt{/ideate} first proposes five candidate directions.
After \texttt{/novelty} checking, candidate A is removed as a duplicate of timing-only feedback approaches, while B, C, D, and E remain for refinement.
The refined candidates are then checked by \texttt{/exp-pilot-run} under the 4$\times$A40 budget: B and C are eliminated because their pilot plans exceed the available cost envelope, D is deferred because its upstream Optimization-Rewind mining would consume the main-run budget, and E is selected for full experimentation.
The final selected path is profiling-guided Claude Code agent optimization, represented as \texttt{claude-code-agent-profiling-guided-gpu}.
The Stage-2 outcomes in the figure are a demonstrative reconstruction projected onto the actual hardware ceiling consumed by the selected idea, rather than a formal pre-experiment screen.
Appendix~\ref{app:bio-idea-pipeline} provides the corresponding pipeline for the biomedical drug discovery case.

\definecolor{cIdea}{HTML}{4A90E2}
\definecolor{cElim}{HTML}{D9534F}
\definecolor{cDefer}{HTML}{B7950B}
\definecolor{cPick}{HTML}{2E8B57}
\definecolor{cStage}{HTML}{6F42C1}
\definecolor{cHW}{HTML}{0E7C7B}

\tikzset{
  stage/.style    ={rectangle, draw=cStage!70, rounded corners=2pt,
                    fill=cStage!12, minimum height=0.7cm, align=center,
                    font=\normalsize\bfseries, inner sep=4pt},
  idea/.style     ={rectangle, draw=cIdea!75, very thick, rounded corners=4pt,
                    fill=cIdea!15, text width=2.7cm, minimum width=2.9cm,
                    minimum height=1.35cm, align=center, font=\footnotesize,
                    inner sep=3pt},
  refined/.style  ={rectangle, draw=cIdea!50, dashed, rounded corners=4pt,
                    fill=white, text width=3.0cm, minimum width=3.25cm,
                    minimum height=1.75cm, align=center, font=\footnotesize,
                    inner sep=3pt},
  elim/.style     ={rectangle, draw=cElim!80, thick, rounded corners=3pt,
                    fill=cElim!15, text width=3.0cm, minimum width=3.25cm,
                    minimum height=1.75cm, align=center, font=\footnotesize,
                    inner sep=3pt},
  defer/.style    ={rectangle, draw=cDefer!85, thick, rounded corners=3pt,
                    fill=cDefer!18, text width=3.0cm, minimum width=3.25cm,
                    minimum height=1.75cm, align=center, font=\footnotesize,
                    inner sep=3pt},
  picked/.style   ={rectangle, draw=cPick!85, very thick, rounded corners=3pt,
                    fill=cPick!22, text width=3.0cm, minimum width=3.25cm,
                    minimum height=1.75cm, align=center, font=\footnotesize,
                    inner sep=3pt},
  hwband/.style   ={rectangle, draw=cHW!70, rounded corners=2pt,
                    fill=cHW!10, minimum height=0.55cm, align=center,
                    font=\footnotesize\itshape, inner sep=4pt},
  ar/.style       ={-{Latex[length=2.5mm]}, semithick, gray!70},
  killar/.style   ={-{Latex[length=2.5mm]}, semithick, dashed, cElim!85},
  passar/.style   ={-{Latex[length=2.5mm]}, semithick, cPick!75}
}

\begin{figure*}[t]
\centering
\resizebox{\textwidth}{!}{%
\begin{tikzpicture}[x=1cm, y=1cm]

  \node[stage, minimum width=18cm] at (0, 0) (S0)
        {Stage 0 \;\textbar\; \texttt{/ideate}
         \; -- direction:\ iterative GPU operator optimization with
         Claude Code based on performance feedback};

  \node[idea] at (-7.4, -1.7) (A)
        {\textbf{A}\\lightweight\\timing-only\\optimizer};
  \node[idea] at (-3.7, -1.7) (B)
        {\textbf{B}\\learned behavioral\\descriptors for\\kernel search};
  \node[idea] at ( 0.0, -1.7) (C)
        {\textbf{C}\\parallel path\\explorer\\(MAP-Elites + agents)};
  \node[idea] at ( 3.7, -1.7) (D)
        {\textbf{D}\\experience-aug.\\iterative kernel\\refinement};
  \node[idea] at ( 7.4, -1.7) (E)
        {\textbf{E}\\profiling-guided\\Claude Code\\agent};

  \foreach \n in {A,B,C,D,E}{\draw[ar] (S0.south -| \n.north) -- (\n.north);}

  \node[stage, minimum width=18cm] at (0, -3.35) (S1)
        {Stage 1 \;\textbar\; \texttt{/novelty}
         \; -- Semantic Scholar + WebSearch + wiki cross-verification};

  \foreach \n in {A,B,C,D,E}{\draw[ar] (\n.south) -- (S1.north -| \n.south);}

  \node[elim]    at (-7.4, -5.2) (Aout)
        {\textbf{A : eliminated}\\
         duplicate approach:\\
         execution timing as\\
         the sole performance\\
         feedback signal};
  \node[refined] at (-3.7, -5.2) (Br)
        {\textbf{B : refined}\\
         CodeT5+ encoder on\\
         MAP-Elites traces;\\
         contrastive vs\\
         hand-crafted 3-D};
  \node[refined] at ( 0.0, -5.2) (Cr)
        {\textbf{C : refined}\\
         30-variant population\\
         + bottleneck-specialised\\
         Verifier agents on a\\
         3-D behavioral grid};
  \node[refined] at ( 3.7, -5.2) (Dr)
        {\textbf{D : refined}\\
         Optimization-Rewind\\
         mining $\to$\\
         retrieval-augmented\\
         refinement loop};
  \node[refined] at ( 7.4, -5.2) (Er)
        {\textbf{E : refined}\\
         Claude Code tool-use:\\
         nsys/ncu $\to$ JSON $\to$\\
         single-hypothesis\\
         targeted edits};

  \draw[killar] (S1.south -| A.south) -- (Aout.north);
  \draw[passar] (S1.south -| B.south) -- (Br.north);
  \draw[passar] (S1.south -| C.south) -- (Cr.north);
  \draw[passar] (S1.south -| D.south) -- (Dr.north);
  \draw[passar] (S1.south -| E.south) -- (Er.north);

  \node[stage,  minimum width=17.2cm] at (0, -7.1) (S2)
        {Stage 2 \;\textbar\; \texttt{/exp-pilot-run}
         \; -- feasibility pilot on the project hardware};
  \node[hwband, minimum width=17.2cm] at (0, -7.8) (HW)
        {Hardware:\ 4$\times$ NVIDIA A40 (sm\_86, 696\,GB/s,
         149.7\,TFLOPS FP16 TC, 48\,GB) \;$\cdot$\;
         Triton 3.2.0 / PyTorch 2.6.0+cu124 \;$\cdot$\;
         pilot budget $\approx$ 250 GPU-hr};

  \draw[passar] (Br.south) -- (Br.south |- S2.north);
  \draw[passar] (Cr.south) -- (Cr.south |- S2.north);
  \draw[passar] (Dr.south) -- (Dr.south |- S2.north);
  \draw[passar] (Er.south) -- (Er.south |- S2.north);

  \node[elim,   minimum height=2.0cm] at (-6.55, -10.1) (Bout)
        {\textbf{B : eliminated (cost)}\\
         MAP-Elites pilot yields\\
         $\sim$120 triples / 8\,GPU-hr;\\
         contrastive needs $\sim$10\,k\\
         samples $\Rightarrow$ \textgreater\,250\\
         GPU-hr just to train\\
         the encoder};
  \node[elim,   minimum height=2.0cm] at (-2.18, -10.1) (Cout)
        {\textbf{C : eliminated (cost)}\\
         30-variant pop. $\times$ per-\\
         variant \texttt{ncu} $\approx$\\
         6\,GPU-hr/op $\times$ 200 ops\\
         $\Rightarrow$ 1.2\,k GPU-hr;\\
         profiler overhead $\sim$3$\times$\\
         A100 on A40};
  \node[defer,  minimum height=2.0cm] at ( 2.18, -10.1) (Dout)
        {\textbf{D : deferred}\\
         Optimization-Rewind\\
         mining at 4--6\,hr/op;\\
         a 200-tuple bank costs\\
         $\sim$1\,k GPU-hr upfront\\
         -- exhausts main-run\\
         budget. Off-deadline};
  \node[picked, minimum height=2.0cm] at ( 6.55, -10.1) (Eout)
        {\textbf{E : pilot passed}\\
         pilot ops\\
          \texttt{dequantize}\\
          \texttt{\_rowwise} +\\
         \texttt{kldiv\_compute}:\\
         5-iter loop $\sim$30\,min/op.\\
         Full sweep $\sim$40\,GPU-hr\\
         -- fits budget};

  \draw[killar] (HW.south -| Bout.north) -- (Bout.north);
  \draw[killar] (HW.south -| Cout.north) -- (Cout.north);
  \draw[killar] (HW.south -| Dout.north) -- (Dout.north);
  \draw[passar] (HW.south -| Eout.north) -- (Eout.north);

  \node[picked, minimum width=11cm, text width=10cm, minimum height=1.1cm,
        font=\normalsize\bfseries] at (0, -13.1) (final)
        {Selected idea \;$\to$\;
         \texttt{claude-code-agent-profiling-guided-gpu}\\[1pt]
         \footnotesize\mdseries
         (NeurIPS 2026 target;\ addresses
         \emph{repair-biased-iterative-refinement} \&
         \emph{correctness-efficiency-gap-kernel-generation})};

  \draw[passar, very thick] (Eout.south) |- (final.east);

\end{tikzpicture}%
}

\caption{Idea screening pipeline for the kernel optimization case.
AutoSci filters candidate directions through novelty checking and pilot experimentation to select one path.}

\label{fig:kernel-idea-pipeline}
\end{figure*}
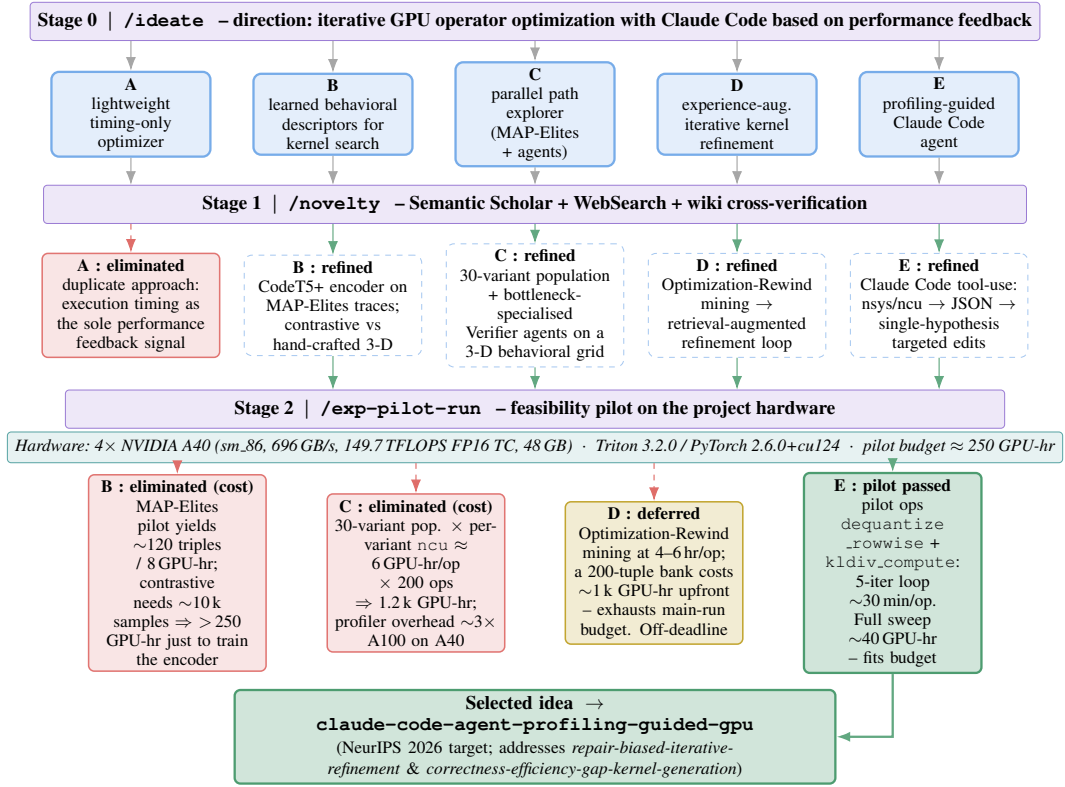
\FloatBarrier

\subsection{Experiment Execution and Analysis}

After selecting the profiling-guided Claude Code agent direction, AutoSci expands the idea into an executable experiment suite rather than a single benchmark run.
Figure~\ref{fig:kernel-exp-taxonomy} summarizes the four experiment blocks.
The sensitivity analysis first fixes the experimental protocol using two reference operators and screens 184 operator prompts into 156 feasible operators.
The main experiment then runs 157 operators for five iterations on the 4$\times$A40 environment; by iteration 5, all 157 generated kernels are executable and correct, with a geometric-mean speedup of 1.52$\times$ over matched baselines, or 1.18$\times$ after excluding degenerate baselines.
The ablation experiment isolates the value of metric feedback by replaying two 60-operator cohorts with a blind autotuning baseline; feedback contributes a 1.58$\times$ gain on the high-headroom cohort and a 1.22$\times$ gain on the broader cohort.
Finally, the intermediate-data analysis audits 628 iteration transitions and shows that most structural changes occur early, while later iterations increasingly become small autotuning adjustments.
The numerical results are reproduced from \texttt{result.md}, \texttt{claude\_ablation\_summary.md}, and \texttt{claude\_iter\_changes.md} in the TritonBench workspace.
Appendix~\ref{app:bio-experiment-suite} provides the corresponding experiment-suite view for the biomedical case, where AutoSci distinguishes executed validation blocks from pre-registered follow-up benchmarks after a negative result.

\definecolor{cMain}{HTML}{1F77B4}
\definecolor{cAbl}{HTML}{D2691E}
\definecolor{cSens}{HTML}{8B5A3C}
\definecolor{cAna}{HTML}{8E44AD}

\tikzset{
  rootexp/.style  ={rectangle, draw=black!60, rounded corners=3pt, very thick,
                    fill=black!8, minimum width=12cm, minimum height=1.0cm,
                    align=center, font=\normalsize\bfseries, inner sep=4pt},
  catBase/.style  ={rectangle, very thick, rounded corners=3pt,
                    text width=3.5cm, minimum width=3.85cm, minimum height=1.0cm,
                    align=center, font=\small\bfseries, inner sep=3pt},
  catSens/.style  ={catBase, draw=cSens!85, fill=cSens!18},
  catMain/.style  ={catBase, draw=cMain!85, fill=cMain!22},
  catAbl/.style   ={catBase, draw=cAbl!85,  fill=cAbl!22},
  catAna/.style   ={catBase, draw=cAna!85,  fill=cAna!18},
  leaf/.style     ={rectangle, draw=black!35, rounded corners=2pt,
                    fill=white, text width=3.6cm, minimum width=3.85cm,
                    minimum height=0.7cm, align=left, font=\footnotesize,
                    inner sep=4pt},
  metric/.style   ={rectangle, draw=black!25, rounded corners=2pt,
                    fill=black!4, text width=3.6cm, minimum width=3.85cm,
                    minimum height=0.7cm, align=left, font=\footnotesize,
                    inner sep=4pt},
}

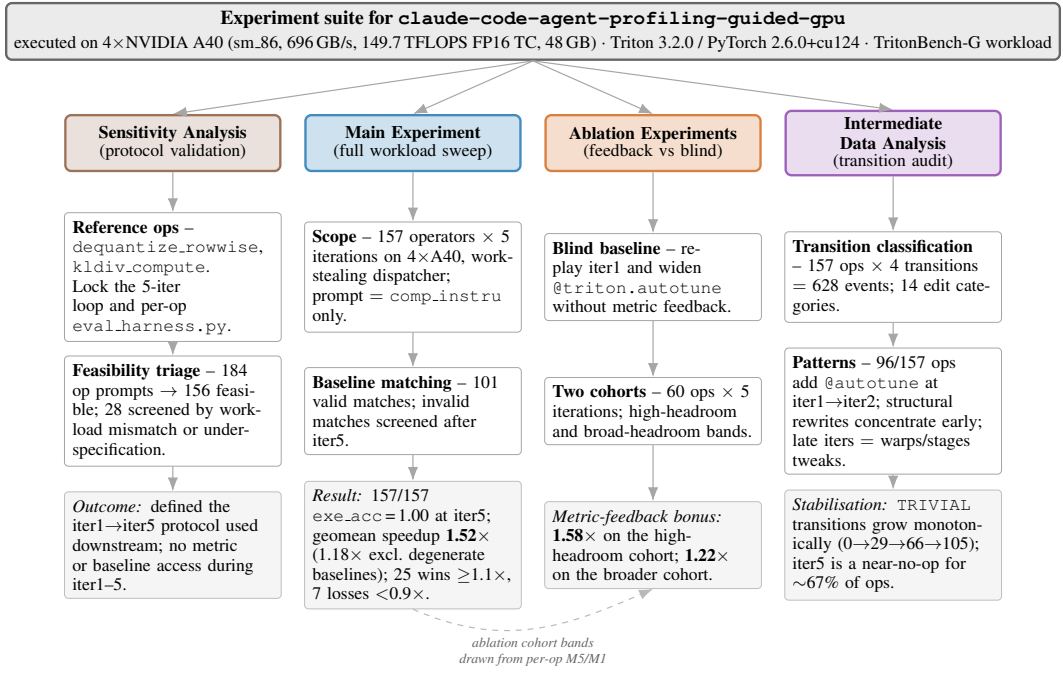
\begin{figure*}[t]
\centering
\resizebox{\textwidth}{!}{%
\begin{tikzpicture}[x=1cm, y=1cm]

  \node[rootexp] at (0, 0) (root)
        {Experiment suite for \texttt{claude-code-agent-profiling-guided-gpu}\\[1pt]
         \footnotesize\mdseries
         executed on 4$\times$NVIDIA A40 (sm\_86, 696\,GB/s,
         149.7\,TFLOPS FP16 TC, 48\,GB) $\cdot$ Triton 3.2.0 /
         PyTorch 2.6.0+cu124 $\cdot$ TritonBench-G workload};

  \node[catSens] at (-6.45, -2.0) (S)
        {Sensitivity Analysis\\[-1pt]
         \footnotesize\mdseries (protocol validation)};
  \node[catMain] at (-2.15, -2.0) (M)
        {Main Experiment\\[-1pt]
         \footnotesize\mdseries (full workload sweep)};
  \node[catAbl]  at ( 2.15, -2.0) (Ab)
        {Ablation Experiments\\[-1pt]
         \footnotesize\mdseries (feedback vs blind)};
  \node[catAna]  at ( 6.45, -2.0) (An)
        {Intermediate Data Analysis\\[-1pt]
         \footnotesize\mdseries (transition audit)};

  \draw[ar] (root.south) -- (S.north);
  \draw[ar] (root.south) -- (M.north);
  \draw[ar] (root.south) -- (Ab.north);
  \draw[ar] (root.south) -- (An.north);

  \node[leaf]   at (-6.45, -4.4) (S1)
        {\textbf{Reference ops} -- \texttt{dequantize\_rowwise},
         \texttt{kldiv\_compute}.\\Lock the 5-iter loop and
         per-op \texttt{eval\_harness.py}.};
  \node[leaf]   at (-6.45, -6.8) (S2)
        {\textbf{Feasibility triage} -- 184 op prompts
         $\to$ 156 feasible; 28 screened by workload mismatch or
         under-specification.};
  \node[metric] at (-6.45, -9.2) (Sm)
        {\emph{Outcome:} defined the iter1$\to$iter5 protocol used
         downstream; no metric or baseline access during iter1--5.};

  \node[leaf]   at (-2.15, -4.4) (M1)
        {\textbf{Scope} -- 157 operators $\times$ 5 iterations on
         4$\times$A40, work-stealing dispatcher;\\prompt $=$
         \texttt{comp\_instru} only.};
  \node[leaf]   at (-2.15, -6.8) (M2)
        {\textbf{Baseline matching} -- 101 valid matches; invalid
         matches screened after iter5.};
  \node[metric] at (-2.15, -9.2) (Mm)
        {\emph{Result:} 157/157 \texttt{exe\_acc}\,=\,1.00 at iter5;\\
         geomean speedup \textbf{1.52$\times$} (1.18$\times$ excl.\
         degenerate baselines); 25 wins $\geq$1.1$\times$,
         7 losses $<$0.9$\times$.};

  \node[leaf]   at ( 2.15, -4.4) (A1)
        {\textbf{Blind baseline} -- replay iter1 and widen
         \texttt{@triton.autotune} without metric feedback.};
  \node[leaf]   at ( 2.15, -6.8) (A2)
        {\textbf{Two cohorts} -- 60 ops $\times$ 5 iterations;
         high-headroom and broad-headroom bands.};
  \node[metric] at ( 2.15, -9.2) (Am)
        {\emph{Metric-feedback bonus:}
         \textbf{1.58$\times$} on the high-headroom cohort;
         \textbf{1.22$\times$} on the broader cohort.};

  \node[leaf]   at ( 6.45, -4.4) (N1)
        {\textbf{Transition classification} --
         157 ops $\times$ 4 transitions $=$ 628 events;
         14 edit categories.};
  \node[leaf]   at ( 6.45, -6.8) (N2)
        {\textbf{Patterns} -- 96/157 ops add \texttt{@autotune}
         at iter1$\to$iter2; structural rewrites concentrate early;
         late iters $=$ warps/stages tweaks.};
  \node[metric] at ( 6.45, -9.2) (Nm)
        {\emph{Stabilisation:} \texttt{TRIVIAL} transitions grow
         monotonically (0$\to$29$\to$66$\to$105); iter5 is a
         near-no-op for $\sim$67\% of ops.};

  \foreach \a/\b in {S/S1, S1/S2, S2/Sm,
                     M/M1, M1/M2, M2/Mm,
                     Ab/A1, A1/A2, A2/Am,
                     An/N1, N1/N2, N2/Nm}{
    \draw[ar] (\a.south) -- (\b.north);
  }

  \draw[ar, dashed, gray!55]
        (Mm.south) .. controls +(1.6,-0.7) and +(-1.6,-0.7) .. (Am.south)
        node[pos=0.5, below=1pt, font=\scriptsize\itshape, color=black!60, align=center]
        {ablation cohort bands\\drawn from per-op M5/M1};

\end{tikzpicture}%
}
\caption{Experiment suite for the selected kernel optimization idea.
AutoSci organizes the selected idea into sensitivity analysis, main evaluation, ablation, and intermediate-data analysis.}
\label{fig:kernel-exp-taxonomy}
\end{figure*}
\FloatBarrier

\subsection{Paper-Level Evaluation}

\begin{table*}[t]
\centering
\footnotesize
\setlength{\tabcolsep}{4pt}
\renewcommand{\arraystretch}{1.18}
\begin{tabularx}{\textwidth}{>{\raggedright\arraybackslash}p{0.17\textwidth}>{\raggedright\arraybackslash}X>{\centering\arraybackslash}p{0.09\textwidth}>{\raggedright\arraybackslash}X}
\hline
\textbf{Case Study} & \textbf{Generated Paper Topic} & \textbf{Score} & \textbf{Review Takeaway} \\
\hline
GPU kernel optimization &
Agent-driven iterative optimization of Triton GPU kernels &
\textbf{6.3 / 10} &
Assessed as a careful empirical study with strong per-iteration traces, controlled ablation, and edit-behavior analysis, but limited by a single model, hardware family, and benchmark suite. \\
Biomedical drug discovery &
PTM-aware degrader target nomination through calibrated ternary-complex scoring &
\textbf{5.8 / 10} &
Assessed as a transparent negative-result paper with useful per-POI calibration and pre-registered follow-up benchmarks, but limited by one main scorer/readout and deferred comparator experiments. \\
\hline
\end{tabularx}
\caption{Automated paper-level review outcomes for the two AutoSci-generated case-study manuscripts. Scores are reported by PaperReview.ai under the ICLR target-venue setting.}
\label{tab:end-to-end-review}
\end{table*}

For each case study, AutoSci runs the \textit{Literature}, \textit{Ideation}, \textit{Experiment}, and \textit{Writing} stages to produce a manuscript-oriented artifact, taking \textbf{27.3} hours for GPU kernel optimization and \textbf{22.6} hours for biomedical drug discovery.
To evaluate whether AutoSci can produce complete paper-level artifacts rather than isolated ideas or experiments, we further conduct an automated review evaluation.
For each case study, AutoSci generates a manuscript-oriented artifact, and we submit the generated paper to PaperReview.ai\footnote{\url{https://paperreview.ai/}} with \emph{ICLR} as the target venue.
The resulting reviews are used as a paper-level review proxy, not as a replacement for formal peer review.
Table~\ref{tab:end-to-end-review} summarizes the reviews.

The reviews suggest that AutoSci can produce reviewable paper-level artifacts rather than only local experimental outputs.
The kernel case is evaluated as a relatively mature empirical paper, while the biomedical case is evaluated as a transparent but incomplete negative-result paper.
Importantly, both reviews expose actionable weaknesses, including limited external validity, missing comparator runs, measurement-noise concerns, and presentation gaps.
These review signals can be stored as submission-stage feedback in SciMem to improve future research workflows.

\section{Related Work}

\subsection{Agent Memory}

Long-horizon agents require memory beyond the fixed context window of an LLM. Prior systems store and retrieve past interactions, user events, or episodic traces to support continuity over time, as in Generative Agents~\citep{park2023generative}, MemoryBank~\citep{zhong2023memorybankenhancinglargelanguage}, and MemGPT~\citep{packerMemGPTLLMsOperating2023}. These works show that memory is a core part of agent behavior rather than a passive cache, especially because longer context alone does not guarantee reliable use of relevant information~\citep{liu2024lost}.
Recent work further moves from flat logs toward structured memory. A-MEM~\citep{xuAMEMAgenticMemory2025} links memory notes in an agentic network, AriGraph~\citep{anokhin2024arigraph} and Zep~\citep{Rasmussen2025Zep} build graph-based world or temporal memories, HippoRAG~\citep{jimenez2024hipporag} combines knowledge graphs with retrieval.

\subsection{Agent Evolution}

Agent evolution has mainly developed along two complementary directions. The first improves an agent through accumulated experience while keeping the base model fixed. Reflexion~\citep{shinn2023reflexion} stores verbal feedback as lessons, ExpeL~\citep{zhao2024expel} extracts reusable experiential knowledge, and Voyager~\citep{wang2023voyager} grows an executable skill library through exploration and feedback. This form of evolution is non-parametric: the agent becomes more capable by reusing memories, strategies, and skills from previous tasks.

The second direction treats the agent system itself as an optimization target. Promptbreeder~\citep{fernando2023promptbreeder} evolves prompts, GPTSwarm~\citep{zhuge2024gptswarm} and AFlow~\citep{zhang2024aflow} search agent graphs or code-represented workflows, and symbolic learning~\citep{zhou2024symbolic} frames prompts, tools, and their composition as learnable symbolic parameters. Recent self-evolving systems further adapt memories, tool libraries, templates, or model behavior, as in SAGE~\citep{liang2024self}, STELLA~\citep{jin2025stella}, and SEAL~\citep{zweiger2025self}.

\section{Conclusion}

In this paper, we presented AutoSci, a memory-centric agentic system for the full scientific research lifecycle.
AutoSci combines schema-governed scientific memory, a harnessed research workflow, DAG-based multi-agent augmentation, and auditable self-evolution.
Together, these modules allow the system to conduct research across literature understanding, ideation, experimentation, writing, and submission feedback while preserving reusable knowledge and experience across projects.
Our case studies in GPU kernel optimization and biomedical drug discovery show that AutoSci can produce reviewable paper-level artifacts from end-to-end research processes.

AutoSci also has important limitations.
First, the current implementation is built as a skill package on top of general-purpose coding and reasoning agents.
This makes the system easy to deploy and inspect, but it is not yet a science-specialized agent foundation.
Future work can develop a research-native agent substrate with models, tools, memory interfaces, and execution protocols designed specifically for scientific work.
Second, evaluation remains underdeveloped.
Existing benchmarks do not yet adequately measure the separate capabilities required by a full research system, including literature understanding, ideation, experimentation, and writing.
We therefore rely on end-to-end case studies and automated review as a paper-level proxy.
A promising direction is to accumulate benchmark tasks from real user workflows, so that future versions of AutoSci can be evaluated both end-to-end and at the level of individual research skills.

\bibliography{ref}
\bibliographystyle{ref}

\clearpage
\appendix
\section{SciDAG Operators and Templates}
\label{app:scidag-operators}

\begin{table}[t]
\centering
\footnotesize
\setlength{\tabcolsep}{4pt}
\renewcommand{\arraystretch}{1.18}
\begin{tabular}{>{\raggedright\arraybackslash}p{0.22\linewidth}>{\raggedright\arraybackslash}p{0.25\linewidth}>{\raggedright\arraybackslash}p{0.43\linewidth}}
\hline
\textbf{Operator} & \textbf{Role} & \textbf{Typical Use} \\
\hline
\texttt{generate} & Exploratory generation & Produce an initial idea, design, or draft candidate. \\
\texttt{variation} & Exploratory diversification & Create alternatives that solve the same task through different assumptions or mechanisms. \\
\texttt{debate} & Multi-view critique & Let specialized agents challenge and refine a candidate through structured disagreement. \\
\texttt{ensemble} & Candidate aggregation & Select or synthesize a stronger output from multiple branches. \\
\texttt{test} & Reliability check & Check code feasibility, experiment design, small-scale execution, or artifact consistency. \\
\texttt{refine} & Test-guided refinement & Revise a candidate according to \texttt{test} feedback before rechecking or downstream use. \\
\texttt{review} & Structured review & Critique a candidate before final editing, focusing on correctness, evidence support, and coherence. \\
\texttt{polish} & Review-guided polish & Improve presentation, wording, formatting, or local organization according to \texttt{review} feedback. \\
\texttt{early-stop} & Control action & Terminate a low-value or already converged branch. \\
\hline
\end{tabular}
\caption{Reusable operators in SciDAG.}
\label{tab:scidag-operators}
\end{table}

\begin{figure*}[t]
\centering
\includegraphics[width=\textwidth]{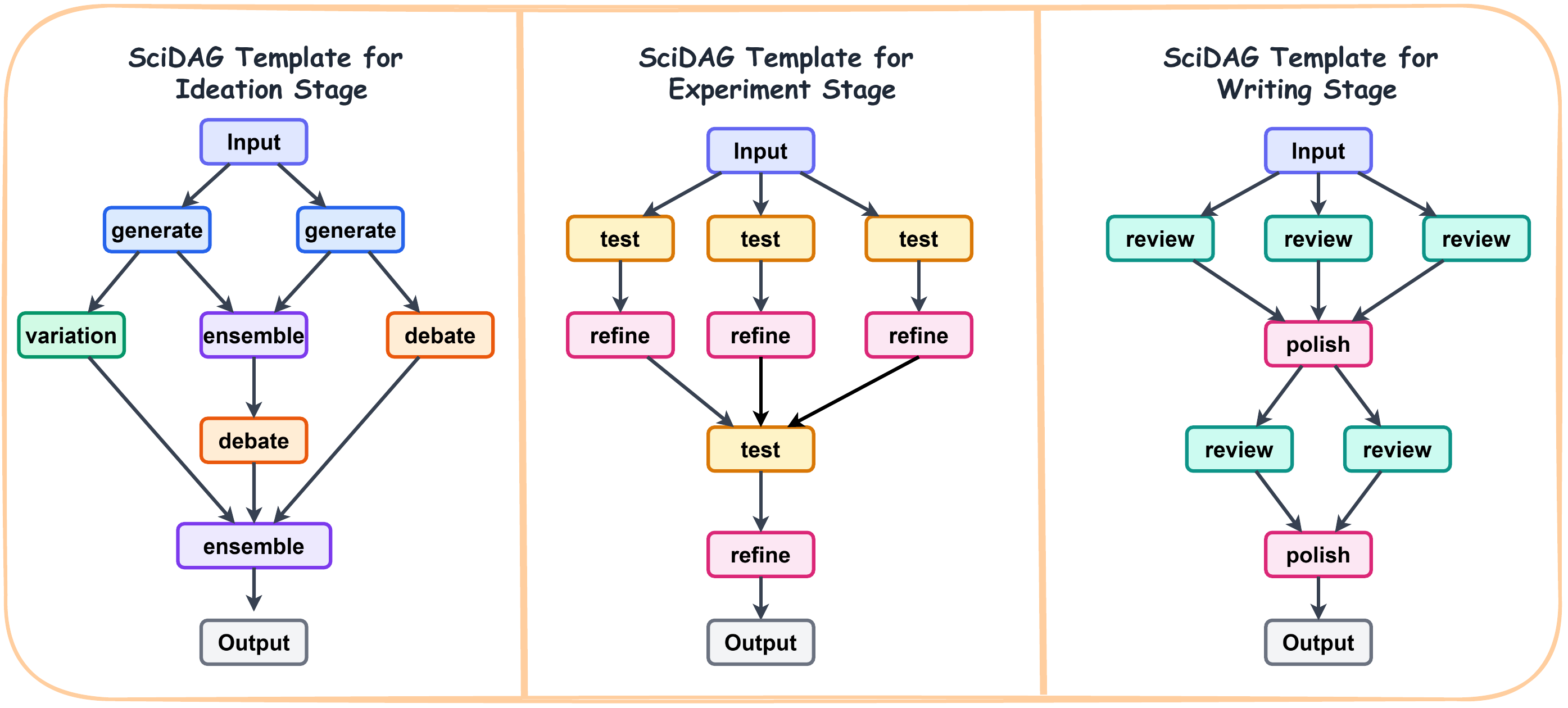}
\caption{Stage-specific SciDAG templates for ideation, experimentation, and writing.}
\label{fig:appendix-scidag-templates}
\end{figure*}

Table~\ref{tab:scidag-operators} lists the reusable operators used by SciDAG, 
and Figure~\ref{fig:appendix-scidag-templates} shows three stage-specific DAG templates for ideation, experimentation, and writing.

\section{Long-Term Knowledge Memory Entity Examples}
\label{app:ltkm-entity-examples}

\definecolor{cConcept}{HTML}{4A90E2}
\definecolor{cTopic}{HTML}{E67E22}
\definecolor{cMethod}{HTML}{2E8B57}
\definecolor{cPaper}{HTML}{6F42C1}
\definecolor{cPeople}{HTML}{1ABC9C}
\definecolor{cFound}{HTML}{D9534F}

\newtcolorbox{wikicard}[2]{
  enhanced,
  breakable,
  colback     = #1!4,
  colframe    = #1!70,
  coltitle    = white,
  title       = \textbf{#2},
  fonttitle   = \sffamily\small,
  boxrule     = 0.6pt,
  arc         = 2mm,
  left        = 6pt, right = 6pt,
  top         = 4pt, bottom = 4pt
}

\newcommand{\kv}[2]{\texttt{\footnotesize #1:}\, {\footnotesize #2}\\}
\newcommand{\bodyhdr}[1]{\smallskip\textit{\#\# #1 \,(excerpt)}\\[1pt]}

The examples below show representative Long-Term Knowledge Memory entity pages generated from the same memory used in Figure~\ref{fig:obsidian-ltkm}.
These examples illustrate how \texttt{Topic}, \texttt{Paper}, \texttt{Foundation}, \texttt{Concept}, \texttt{Method}, and \texttt{People} entities are stored as schema-governed pages with frontmatter, body sections, and typed cross-references.

\begin{wikicard}{cConcept}{concept \,$|$\, \texttt{repair-biased-iterative-refinement}}
\footnotesize
\textit{Frontmatter (selected)}\\[1pt]
\kv{title}{Repair-Biased Iterative Refinement}
\kv{aliases}{%
  \texttt{[repair-vs-optimization-iteration,}\\
  \texttt{iterative-refinement-repair-bias]}%
  }
\kv{tags}{\texttt{[llm-code-generation, gpu-kernels, iterative-refinement,
                  agent-systems]}}
\kv{maturity}{emerging}
\kv{first\_introduced}{2026-05}
\kv{key\_papers}{\texttt{kernelbenchx-benchmark-evaluating-llm-generated}}
\kv{related\_concepts}{\texttt{iterative-kernel-refinement},
                       \texttt{correctness-efficiency-gap-kernel-generation}}
\dots \\

\bodyhdr{Definition}
The empirical observation that iterative refinement loops in LLM-based
kernel generation systems (such as GEAK) operate primarily as \emph{repair} mechanisms rather than \emph{optimization} mechanisms.
Across GEAK iterations, compile success rises from 52.3\% to 68.8\% and
correctness from 18.2\% to 30.7\%, but average speedup falls from
1.58$\times$ to 1.44$\times$\,\dots

\bodyhdr{Intuition}
The underlying asymmetry is structural: repair responds to explicit,
local error signals (compile errors, shape mismatches), while
performance improvement requires plan-level decisions about tiling,
memory layout and kernel boundaries -- decisions not recoverable from
the feedback available in current iterative pipelines\,\dots

\smallskip
\textit{\#\# ...}
\end{wikicard}
\noindent\textit{\textbf{Concept entry.} A stable named idea synthesised from
one or more papers. Lives in \texttt{wiki/concepts/}; the
\texttt{related\_concepts} and \texttt{key\_papers} fields drive most of
the in-graph edges to other concepts and to papers.}
\medskip

\begin{wikicard}{cTopic}{topic \,$|$\, \texttt{llm-kernel-generation}}
\footnotesize
\textit{Frontmatter (selected)}\\[1pt]
\kv{title}{LLM-Based Kernel Generation}
\kv{tags}{\texttt{[llm, kernel-generation, gpu, code-generation, compiler]}}
\kv{related\_topics}{\texttt{[]}\, (none yet)}
\kv{key\_people}{\texttt{[]}\, (populated via reverse-xref from people pages)}
\kv{linked\_ideas}{\texttt{[]}\, (populated when an idea declares
                                 \texttt{origin\_gaps} pointing here)}
\dots \\

\bodyhdr{Overview}
LLM-based kernel generation is the task of using language models to
automatically produce high-performance GPU or accelerator kernels
from high-level specifications. It combines code generation with
hardware-aware optimization.

\bodyhdr{SOTA tracker}
\begin{tabular}{@{}p{3.8cm}p{3.2cm}p{6.0cm}@{}}
\textbf{Method} & \textbf{Benchmark} & \textbf{Correctness / Speedup} \\[2pt]
Frontier reasoning models   & KernelBench     & $<$20\% match PyTorch baseline \\
\texttt{kernelfoundry-...}  & KernelBench     & 97\% (SYCL), 2.32$\times$ avg  \\
AscendOptimizer             & 101 AscendC ops & 1.21$\times$ geo-mean         \\
\end{tabular}

\bodyhdr{Open problems}
Fusion tasks: 72\% failure rate across all methods (KernelBenchX).
Correctness vs. efficiency disconnect: 46.6\% of correct kernels are
\emph{slower} than baseline. Cross-hardware variance: speedup ratio
reaches 21.4$\times$ across NVIDIA GPUs\,\dots

\smallskip\textit{\#\# ... \,}\\
\end{wikicard}
\noindent\textit{\textbf{Topic entry.} A research area that aggregates many
concepts, methods and papers. Lives in \texttt{wiki/topics/}. The body
sections (\texttt{Timeline}, \texttt{Seminal works}, \texttt{SOTA
tracker}, \texttt{Open problems}) make a topic page the natural landing
point for new readers and the place where cross-paper synthesis is
recorded.}
\medskip

\begin{wikicard}{cMethod}{method \,$|$\, \texttt{optimization-rewind}}
\footnotesize
\textit{Frontmatter (selected)}\\[1pt]
\kv{name}{Optimization Rewind}
\kv{type}{optimization}
\kv{tags}{\texttt{[self-supervised, kernel-optimization,
                  experience-mining, de-optimization,
                  knowledge-scarcity]}}
\kv{source\_papers}{\texttt{ascendoptimizer-episodic-agent-ascend-npu-operator}}
\kv{parent\_methods}{\texttt{[]}}\kv{child\_methods}{\texttt{[]}}
\kv{code\_repo}{\texttt{https://github.com/KernelHive}}
\kv{date\_updated}{2026-05-24}
\dots \\

\bodyhdr{Mechanism}
A self-supervised technique for mining reusable optimization
knowledge from existing high-performance implementations. It works
by systematically removing recognizable optimization motifs from
strong kernels and retaining only the removals that measurably
degrade hardware performance\,\dots

\bodyhdr{Procedure}
\begin{enumerate}\setlength{\itemsep}{0pt}\setlength{\topsep}{1pt}
\item Start from the strongest available implementation of an operator.
\item An inverse agent proposes semantically meaningful
      de-optimizations (removing pipelining, breaking vectorized
      paths, reintroducing synchronization, \dots).
\item Each de-optimized candidate is compiled, correctness-checked
      and profiled on real hardware.
\item Validated degradations are distilled into experience tuples
      \texttt{(Title, Bottleneck, Applicability, Effect, Diff)}.
\item \dots
\end{enumerate}

\bodyhdr{Assumptions}
Optimization motifs are \emph{compositional} -- they can be removed
and added independently, so a single-factor screen
($\Delta_m = \mathrm{Lat}(K_{A\setminus\{m\}}) - \mathrm{Lat}(K_A)$)
recovers each motif's marginal contribution. The profiling noise
threshold $\tau_{\text{noise}}$ is taken to be stable across
de-optimizations\,\dots

\smallskip\textit{\#\# ... \,}\\
\end{wikicard}
\noindent\textit{\textbf{Method entry.} A concrete system, algorithm or
benchmark that realises one or more concepts. Lives in
\texttt{wiki/methods/}. \texttt{source\_papers} pins it to its paper of
record; \texttt{parent\_methods} / \texttt{child\_methods} capture
method lineage when a system extends or specialises another.}
\medskip

\begin{wikicard}{cPaper}{paper \,$|$\, \texttt{kernelbench-llms-write-efficient-gpu-kernels}}
\footnotesize
\textit{Frontmatter (selected)}\\[1pt]
\kv{title}{KernelBench: Can LLMs Write Efficient GPU Kernels?}
\kv{arxiv}{2502.10517}\kv{year}{2025}\kv{importance}{4}
\kv{source\_type}{tex}\kv{date\_added}{2025-05-24}
\kv{contribution\_type}{\texttt{[benchmark, method]}}
\kv{datasets}{\texttt{[KernelBench]}}
\kv{tldr}{Introduces KernelBench, a benchmark of 250 PyTorch workloads
          showing frontier reasoning models match the PyTorch baseline
          on \textless20\% of tasks.}
\dots \\

\bodyhdr{Key idea}
A standardised evaluation framework with 250 PyTorch workloads at three
difficulty levels (single ops, op sequences, full architectures). The
\texttt{fast\_p} metric -- percentage of kernels both correct \emph{and}
faster than baseline by factor $p$ -- captures correctness and
performance in a single adjustable-threshold measure\,\dots

\bodyhdr{Experiment \& Results}
One-shot \texttt{fast\_1} over PyTorch eager: Level~1: DeepSeek-R1
12\% (others 3--10\%); Level~2: DeepSeek-R1 36\% (others 0--24\%);
Level~3: OpenAI o1 12\% (others 0--8\%). Iterative refinement (10
turns) lifts DeepSeek-R1 Level~2 from 36\% to 72\%\,\dots

\bodyhdr{Limitations}
Limited to NVIDIA GPUs; CUDA is low-resource in pre-training
($\approx$0.07\% of The Stack v1.2); models rarely use tensor cores /
wmma; functional-correctness errors are hard to fix via feedback;
few-shot examples \emph{degrade} \texttt{fast\_1} by encouraging
aggressive but error-prone optimisations\,\dots

\smallskip\textit{\#\# ... \,}\\
\end{wikicard}
\noindent\textit{\textbf{Paper entry.} A first-class citizen -- each paper has
its own page rather than only an entry in a bibliography. Lives in
\texttt{wiki/papers/}. \texttt{importance} (1--5) drives sort order in
the index; \texttt{contribution\_type} lets \texttt{/discover} and
\texttt{/survey} retrieve ``benchmark papers'' vs. ``method papers''
without re-reading the bodies.}
\medskip

\begin{wikicard}{cPeople}{people \,$|$\, \texttt{jun-zhu}}
\footnotesize
\textit{Frontmatter (selected)}\\[1pt]
\kv{name}{Jun Zhu}
\kv{affiliation}{\textit{(unset)}}
\kv{research\_areas}{\texttt{[llm-code-generation, gpu-kernels, systems]}}
\kv{type.kind}{researcher}
\kv{homepage}{\textit{(unset)}}\kv{scholar}{\textit{(unset)}}
\kv{date\_updated}{2026-05-24}
\dots \\

\bodyhdr{Research areas}
LLM-based code generation, GPU kernel generation, systems\,\dots

\bodyhdr{Recent work}
\begin{itemize}\setlength{\itemsep}{0pt}\setlength{\topsep}{1pt}
\item \texttt{[[kernelbenchx-comprehensive-benchmark-evaluating-llm-generated]]}
\item \dots
\end{itemize}
\textit{(populated by the \texttt{papers $\to$ people} reverse-xref rule
whenever a paper body wikilinks this author slug.)}\\

\smallskip\textit{\#\# ... \,}\\

\end{wikicard}
\noindent\textit{\textbf{People entry.} Lightweight by design -- only the
fields needed to attribute and disambiguate. Lives in
\texttt{wiki/people/}. The \texttt{Recent work} section is populated
automatically by the \texttt{papers $\to$ people} reverse-xref rule, so
the page stays in sync with the literature without manual edits.}
\medskip

\begin{wikicard}{cFound}{foundation \,$|$\, \texttt{operator-fusion}}
\footnotesize
\textit{Frontmatter (selected)}\\[1pt]
\kv{title}{Operator Fusion}
\kv{domain}{compilers and high-performance computing}
\kv{status}{mainstream}
\kv{aliases}{\texttt{[kernel fusion, op fusion, loop fusion]}}
\kv{first\_introduced}{1980s loop fusion; revived in TVM (2018) / XLA (2017)}
\kv{date\_updated}{2026-05-27}
\kv{source\_url}{\textit{(unset)}}
\dots \\

\bodyhdr{Definition}
The program-transformation technique of combining the bodies of two or
more separately specified operators into a single executable kernel,
so that intermediate values flowing between them are \emph{not}
materialised in main memory. Semantics are preserved; only the schedule
changes\,\dots

\bodyhdr{Intuition}
GPU kernels are bandwidth-bound: moving an element from DRAM to a
register dwarfs the arithmetic on it. Fusing \texttt{y=f(x)} and
\texttt{z=g(y)} into \texttt{z=g(f(x))} keeps \texttt{y} on-chip, killing
one DRAM round-trip per element \emph{and} one kernel-launch overhead
per call -- pushing the kernel from bandwidth-bound toward
compute-bound\,\dots

\bodyhdr{Relevance to active research}
A load-bearing assumption of every downstream concept in this wiki:
\texttt{[[gpu-kernel-generation]]}, \texttt{[[iterative-kernel-refinement]]},
and the \texttt{[[correctness-efficiency-gap-kernel-generation]]} all
sit on top of ``good fusion is the primary lever for GPU performance.''
FlashAttention is a tour-de-force application; KernelBench Level~2 is
essentially a fusion benchmark\,\dots

\smallskip\textit{\#\# ... \,}\\
\end{wikicard}
\noindent\textit{\textbf{Foundation entry.} Textbook background that newer
research builds on, separated from active-research \texttt{concepts/} so
the rest of the wiki can wikilink to ``well-known prerequisites'' without
inflating the concept list. Lives in \texttt{wiki/foundations/}.}

\section{Biomedical Drug Discovery Idea Pipeline}
\label{app:bio-idea-pipeline}

Figure~\ref{fig:bio-idea-pipeline} shows the idea screening and lifecycle pipeline for the biomedical drug discovery case study.
Unlike the kernel case in the main text, this case produces a negative result that becomes useful: the refuted premise and a deferred sibling idea jointly define the next feasible research direction.
Specifically, \texttt{/ideate} first proposes five candidate directions for structure-aware PTM modelling.
After \texttt{/novelty} checking, A is removed because the PTM-site disorder-prediction subspace is already saturated, and B is removed because the required AF3 fine-tuning is infeasible under the available constraints.
C, D, and E remain as refined candidates.
The \texttt{/exp-design} stage then prioritizes these survivors: C and D are deferred, while E, PTM-aware degrader target nomination, is selected for execution because its Phase-0 floor test is cheap, fast, and feasible on the available RTX 4060 environment.
The selected idea is decomposed into two sub-claims and evaluated with real operator runs.
The result refutes the load-bearing premise under current PTM-blind scorers, and the post-mortem combines this negative evidence with the deferred PTM-conditioning idea to regenerate a feasible follow-up plan.

\definecolor{cFail}{HTML}{8B1A1A}
\definecolor{cRegen}{HTML}{138D75}
\definecolor{cDecomp}{HTML}{5DADE2}

\providecommand{\dpt}{$\Delta$pTernary}

\tikzset{
  stage/.style    ={rectangle, draw=cStage!70, rounded corners=2pt,
                    fill=cStage!12, minimum height=0.7cm, align=center,
                    font=\normalsize\bfseries, inner sep=4pt},
  idea/.style     ={rectangle, draw=cIdea!75, very thick, rounded corners=4pt,
                    fill=cIdea!15, text width=2.7cm, minimum width=2.9cm,
                    minimum height=1.35cm, align=center, font=\footnotesize,
                    inner sep=3pt},
  refined/.style  ={rectangle, draw=cIdea!50, dashed, rounded corners=4pt,
                    fill=white, text width=3.0cm, minimum width=3.25cm,
                    minimum height=1.9cm, align=center, font=\footnotesize,
                    inner sep=3pt},
  elim/.style     ={rectangle, draw=cElim!80, thick, rounded corners=3pt,
                    fill=cElim!15, text width=3.0cm, minimum width=3.25cm,
                    minimum height=1.9cm, align=center, font=\footnotesize,
                    inner sep=3pt},
  defer/.style    ={rectangle, draw=cDefer!85, thick, rounded corners=3pt,
                    fill=cDefer!18, text width=3.0cm, minimum width=3.25cm,
                    minimum height=1.9cm, align=center, font=\footnotesize,
                    inner sep=3pt},
  picked/.style   ={rectangle, draw=cPick!85, very thick, rounded corners=3pt,
                    fill=cPick!22, text width=3.0cm, minimum width=3.25cm,
                    minimum height=1.9cm, align=center, font=\footnotesize,
                    inner sep=3pt},
  decomp/.style   ={rectangle, draw=cDecomp!85, thick, rounded corners=3pt,
                    fill=cDecomp!15, text width=3.2cm, minimum width=3.4cm,
                    minimum height=1.25cm, align=center, font=\footnotesize,
                    inner sep=3pt},
  failbox/.style  ={rectangle, draw=cFail!85, very thick, rounded corners=3pt,
                    fill=cFail!12, text width=10.5cm, minimum width=11cm,
                    minimum height=1.4cm, align=center, font=\footnotesize,
                    inner sep=4pt},
  postm/.style    ={rectangle, draw=cStage!70, thick, rounded corners=3pt,
                    fill=cStage!10, text width=6.2cm, minimum width=6.5cm,
                    minimum height=1.9cm, align=center, font=\footnotesize,
                    inner sep=4pt},
  regen/.style    ={rectangle, draw=cRegen!85, very thick, rounded corners=3pt,
                    fill=cRegen!16, text width=6.2cm, minimum width=6.5cm,
                    minimum height=1.9cm, align=center, font=\footnotesize,
                    inner sep=4pt},
  dchip/.style    ={rectangle, draw=cDefer!85, thick, dashed, rounded corners=3pt,
                    fill=cDefer!16, text width=3.2cm, minimum width=3.4cm,
                    minimum height=1.6cm, align=center, font=\footnotesize,
                    inner sep=3pt},
  hwband/.style   ={rectangle, draw=cHW!70, rounded corners=2pt,
                    fill=cHW!10, minimum height=0.55cm, align=center,
                    font=\footnotesize\itshape, inner sep=4pt},
  ar/.style       ={-{Latex[length=2.5mm]}, semithick, gray!70},
  killar/.style   ={-{Latex[length=2.5mm]}, semithick, dashed, cElim!85},
  passar/.style   ={-{Latex[length=2.5mm]}, semithick, cPick!75},
  regar/.style    ={-{Latex[length=2.5mm]}, semithick, cRegen!80},
  weakar/.style   ={-{Latex[length=2.0mm]}, thin, dashed, gray!55}
}

\begin{figure*}[p]
\centering
\resizebox{\textwidth}{!}{%
\begin{tikzpicture}[x=1cm, y=1cm]

  \node[stage, minimum width=20cm] at (0, 0) (S0)
        {Stage 0 \;\textbar\; \texttt{/ideate}
         \; -- direction:\ structure-aware PTM modelling for drug discovery};

  \node[idea] at (-8.0, -2.0) (A)
        {\textbf{A}\\PTM-site\\disorder predictor\\(pLDDT + IDR)};
  \node[idea] at (-4.0, -2.0) (B)
        {\textbf{B}\\chirality-aware\\AF3 diffusion\\noise schedule};
  \node[idea] at ( 0.0, -2.0) (C)
        {\textbf{C}\\PTM-resolved\\structural\\interactome\\($\Delta$pDockQ-per-PTM)};
  \node[idea] at ( 4.0, -2.0) (D)
        {\textbf{D}\\PTM-conditioned\\ensemble (pair-rep\\scaling adapter)};
  \node[idea] at ( 8.0, -2.0) (E)
        {\textbf{E}\\PTM-aware degrader\\target nomination\\(\dpt)};

  \foreach \n in {A,B,C,D,E}{\draw[ar] (S0.south -| \n.north) -- (\n.north);}

  \node[stage, minimum width=20cm] at (0, -4.0) (S1)
        {Stage 1 \;\textbar\; \texttt{/novelty}
         \; -- Semantic Scholar + WebSearch + PubMed + wiki cross-verification};

  \foreach \n in {A,B,C,D,E}{\draw[ar] (\n.south) -- (S1.north -| \n.south);}

  \node[elim]    at (-8.0, -6.4) (Aout)
        {\textbf{A : eliminated}\\
         subspace saturated:\\
         SAPP / PhosAF /\\
         GraphPhos / AstraPTM2 /\\
         DeepPCT / MTPrompt-PTM\\
         ($\geq$5 2024--25 entries)};
  \node[elim]    at (-4.0, -6.4) (Bout)
        {\textbf{B : eliminated}\\
         not feasible:\\
         AF3 weights are\\
         non-commercial; an\\
         external group cannot\\
         fine-tune the diffusion head};
  \node[refined] at ( 0.0, -6.4) (Cr)
        {\textbf{C : refined}\\
         restrict to PTM-near-\\
         interface edges; pre-\\
         register 4-case holdout\\
         (14-3-3/Cdc25C, HIF1$\alpha$/\\
         pVHL, PCNA-K164, FOXO3a)};
  \node[refined] at ( 4.0, -6.4) (Dr)
        {\textbf{D : refined}\\
         PTM adapter on a\\
         frozen Boltz-2 backbone;\\
         train to ensemble\\
         populations; Boltz-sample\\
         is the head-to-head baseline};
  \node[refined] at ( 8.0, -6.4) (Er)
        {\textbf{E : refined}\\
         per-POI noise-floor\\
         calibration as the\\
         load-bearing fail-fast\\
         gate; MD-relaxed route\\
         decouples from a PTM-Cfold};

  \draw[killar] (S1.south -| A.south) -- (Aout.north);
  \draw[killar] (S1.south -| B.south) -- (Bout.north);
  \draw[passar] (S1.south -| C.south) -- (Cr.north);
  \draw[passar] (S1.south -| D.south) -- (Dr.north);
  \draw[passar] (S1.south -| E.south) -- (Er.north);

  \node[stage,  minimum width=13cm] at (4.0, -8.7) (S2)
        {Stage 2 \;\textbar\; \texttt{/exp-design}
         \; -- composite-score prioritisation of the three survivors};

  \draw[passar] (Cr.south) -- (Cr.south |- S2.north);
  \draw[passar] (Dr.south) -- (Dr.south |- S2.north);
  \draw[passar] (Er.south) -- (Er.south |- S2.north);

  \node[defer,  minimum height=2.0cm] at ( 0.0, -11.1) (Cdef)
        {\textbf{C : deferred}\\
         composite 10. Proteome-\\
         scale AF2-Multimer fold\\
         is off-budget; AF3-phospho\\
         collapse risk on the\\
         conformational cases};
  \node[defer,  minimum height=2.0cm] at ( 4.0, -11.1) (Ddef)
        {\textbf{D : deferred}\\
         composite 13. Scoop risk:\\
         Boltz-sample (Jan 2026)\\
         already does PTM-blind\\
         pair-rep scaling; needs an\\
         ensemble-labelled trainset};
  \node[picked, minimum height=2.0cm] at ( 8.0, -11.1) (Epick)
        {\textbf{E : selected}\\
         composite 16, priority 5.\\
         Phase-0 floor is cheap and\\
         kills-or-de-risks fast;\\
         public weights, GREEN\\
         feasibility on an RTX 4060};

  \draw[weakar] (Cr.south |- S2.south) -- (Cdef.north);
  \draw[weakar] (Dr.south |- S2.south) -- (Ddef.north);
  \draw[passar] (Er.south |- S2.south) -- (Epick.north);

  \node[stage,  minimum width=16.5cm] at (2.0, -13.5) (S3)
        {Stage 3 \;\textbar\; \texttt{/exp-design $\to$ /exp-run $\to$ /exp-eval}
         \; -- selected idea decomposed into 2 sub-claims and executed};
  \node[hwband, minimum width=16.5cm] at (2.0, -14.3) (HW)
        {Real blocks executed on 1$\times$ NVIDIA RTX 4060 (8\,GB)
         \;$\cdot$\; DeepTernary v1.0.0 + PROTAC-STAN inference repos
         \;$\cdot$\; pre-registered blocks frozen, not executed};

  \draw[passar] (Epick.south) -- (Epick.south |- S3.north);

  \node[decomp] at (-1.0, -16.2) (sub1)
        {sub-claim 1\\\textbf{noise-floor-calibrated\\\dpt\ improves ranking}};
  \node[decomp] at ( 5.0, -16.2) (sub2)
        {sub-claim 2\\\textbf{MD-relaxed phospho route\\$\approx$ native CCD-PTM token}\\\textit{\footnotesize(decouples from a PTM-Cfold, cf.\ idea D)}};

  \draw[ar] (HW.south -| sub1.north) -- (sub1.north);
  \draw[ar] (HW.south -| sub2.north) -- (sub2.north);

  \node[failbox] at (2.0, -18.9) (neg)
        {\textbf{Real-operator pilot $\Rightarrow$ load-bearing premise REFUTED}
         (bounded to current PTM-blind scorers).\\
         15 POIs / 189 interface sites: phospho 14.5\%, alanine 15.9\%,
         Kme3 15.7\% clear the per-POI floor vs 13.4\% chance
         ($p>0.3$; effect-size CIs include 0; \textbf{0/69} survive
         BH-FDR / Bonferroni). Dose-response control proves the operator is
         \emph{non-inert} $\to$ the bottleneck is the scorer's dynamic range.};

  \draw[killar] (sub1.south) -- (sub1.south |- neg.north);
  \draw[killar] (sub2.south) -- (sub2.south |- neg.north);

  \node[postm,  minimum width=6.0cm, text width=5.7cm] at (-4.0, -21.4) (pm)
        {\textbf{Post-mortem}\\
         the limit is the structure$\to$score chain, not the readout or the
         threshold: a PTM-\emph{blind} scorer cannot be wrapped into a
         PTM-selective one. Null-matching must use the identical operation
         as the signal.};
  \node[dchip] at ( 3.5, -21.4) (dchip)
        {\textbf{idea D re-enters}\\(deferred)\\PTM-conditioning supplies\\the missing scorer};

  \draw[killar] (neg.south west) -- (pm.north);

  \node[regen,  minimum width=7.4cm, text width=7.1cm] at (0.8, -23.8) (rg)
        {\textbf{Regenerated feasible plan} (new idea from two)\\
         next-gen idea $=$ a \emph{PTM-sensitive} ternary scorer --- the
         explicit prerequisite. Deliverable $=$ the negative bound $+$ the
         frozen-threshold benchmark a future scorer must clear.};

  \draw[regar]        (pm.south)    -- (rg.north west);
  \draw[regar,dashed] (dchip.south) -- (rg.north east);

\end{tikzpicture}%
}
\caption{Idea screening and lifecycle pipeline for the biomedical case.
AutoSci filters candidate directions, executes selected paths, and converts negative evidence into a regenerated follow-up plan.}
\label{fig:bio-idea-pipeline}
\end{figure*}
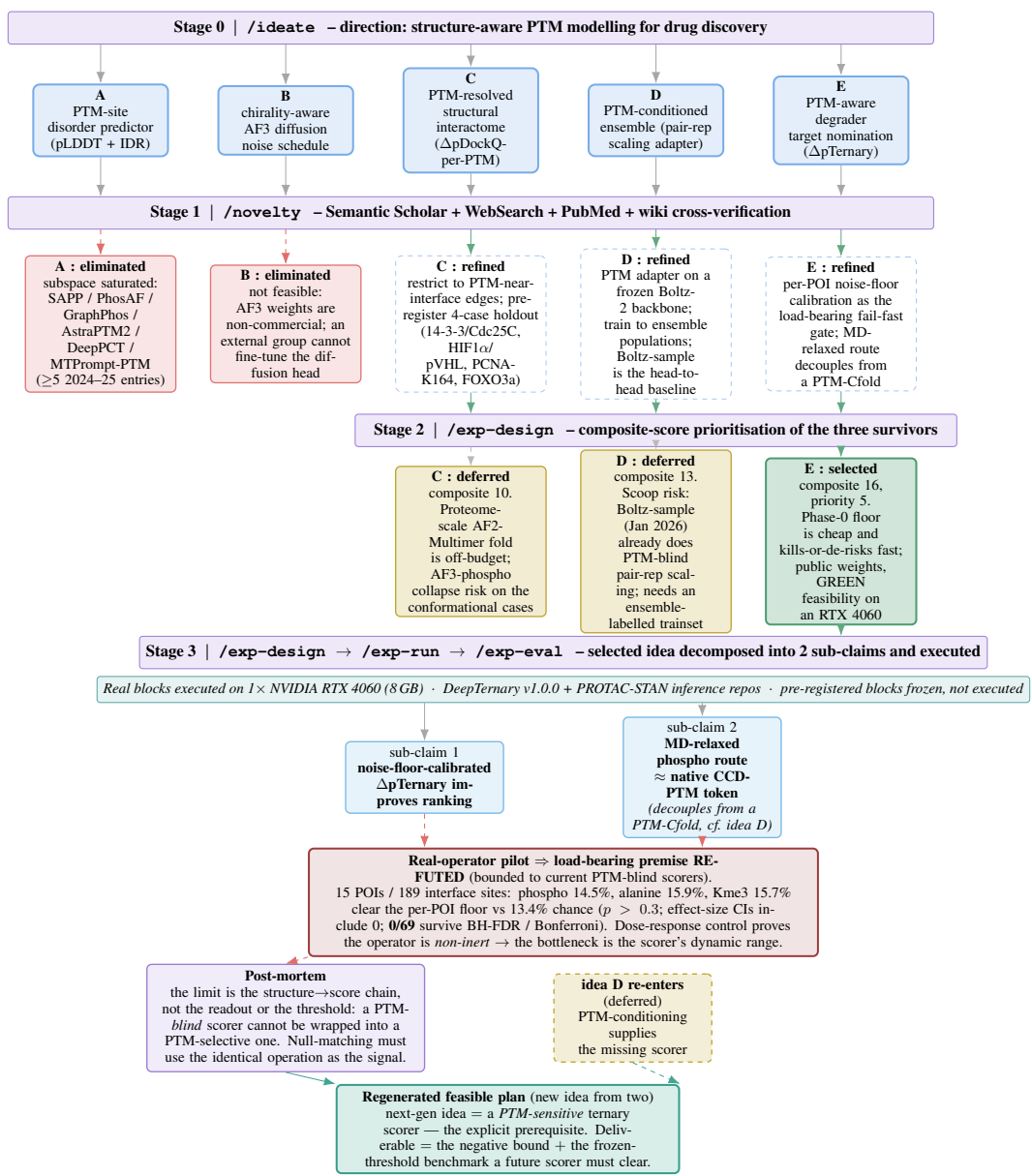
\FloatBarrier

\section{Biomedical Drug Discovery Experiment Suite}
\label{app:bio-experiment-suite}

Figure~\ref{fig:bio-exp-taxonomy} shows the experiment suite for the selected biomedical idea, \texttt{ptm-aware-degrader-target-nomination}.
The suite separates executed experiments from frozen follow-up benchmarks.
The executed blocks reproduce the base DeepTernary scorer, calibrate a per-protein noise floor, and test whether PTM-like interface perturbations produce signal beyond that floor.
The result is negative: the perturbations do not exceed chance-level clearance under the current PTM-blind scorer, and no site survives multiplicity correction.
Rather than discarding this outcome, AutoSci turns the negative bound into a pre-registered benchmark that a future PTM-sensitive scorer must clear.

\definecolor{cBioMain}{HTML}{1F77B4}
\definecolor{cBioSens}{HTML}{8B5A3C}
\definecolor{cBioCore}{HTML}{8B1A1A}
\definecolor{cBioFroz}{HTML}{B7950B}
\definecolor{cBioFail}{HTML}{8B1A1A}

\providecommand{\dpt}{$\Delta$pTernary}

\tikzset{
  bioRootExp/.style ={rectangle, draw=black!60, rounded corners=3pt, very thick,
                    fill=black!8, minimum width=13.8cm, minimum height=1.0cm,
                    align=center, font=\normalsize\bfseries, inner sep=4pt},
  bioCatBase/.style ={rectangle, very thick, rounded corners=3pt,
                    text width=3.7cm, minimum width=4.0cm, minimum height=1.0cm,
                    align=center, font=\small\bfseries, inner sep=3pt},
  bioCatMain/.style ={bioCatBase, draw=cBioMain!85, fill=cBioMain!22},
  bioCatSens/.style ={bioCatBase, draw=cBioSens!85, fill=cBioSens!18},
  bioCatCore/.style ={bioCatBase, draw=cBioCore!85, fill=cBioCore!18},
  bioCatFroz/.style ={bioCatBase, draw=cBioFroz!85, fill=cBioFroz!18},
  bioLeaf/.style    ={rectangle, draw=black!35, rounded corners=2pt,
                    fill=white, text width=3.7cm, minimum width=4.0cm,
                    minimum height=0.72cm, align=left, font=\footnotesize,
                    inner sep=4pt},
  bioMetric/.style  ={rectangle, draw=black!25, rounded corners=2pt,
                    fill=black!4, text width=3.7cm, minimum width=4.0cm,
                    minimum height=0.72cm, align=left, font=\footnotesize,
                    inner sep=4pt},
  bioMetricNeg/.style={bioMetric, draw=cBioFail!65, fill=cBioFail!10},
}

\begin{figure}[!h]
\centering
\resizebox{\textwidth}{!}{%
\begin{tikzpicture}[x=1cm, y=1cm]

  \node[bioRootExp] at (0, 0) (root)
        {Experiment suite for \texttt{ptm-aware-degrader-target-nomination}\\[1pt]
         \footnotesize\mdseries
         real blocks on 1$\times$NVIDIA RTX 4060 (8\,GB) $\cdot$ DeepTernary
         v1.0.0 + PROTAC-STAN $\cdot$ solid = executed, dashed = frozen benchmark};

  \node[bioCatMain] at (-7.05, -2.1) (P)
        {Precondition / Baseline\\[-1pt]
         \footnotesize\mdseries (scorer reproduction)};
  \node[bioCatSens] at (-2.35, -2.1) (Ca)
        {Calibration / Sensitivity\\[-1pt]
         \footnotesize\mdseries (per-POI noise floor)};
  \node[bioCatCore] at ( 2.35, -2.1) (Co)
        {Core Operator\\[-1pt]
         \footnotesize\mdseries (primary negative finding)};
  \node[bioCatFroz] at ( 7.05, -2.1) (F)
        {Pre-registered Benchmark\\[-1pt]
         \footnotesize\mdseries (frozen follow-up)};

  \draw[ar] (root.south) -- (P.north);
  \draw[ar] (root.south) -- (Ca.north);
  \draw[ar] (root.south) -- (Co.north);
  \draw[ar] (root.south) -- (F.north);

  \node[bioLeaf]   at (-7.05, -4.6) (P1)
        {\textbf{DeepTernary reproduction} -- v1.0.0 on the
         22-complex unbound PROTAC test set.};
  \node[bioLeaf]   at (-7.05, -6.7) (P2)
        {\textbf{Reproduction metric} -- DockQ top-1
         \textbf{0.397} vs authors' 0.429; Acceptable-Rate 1.0.};
  \node[bioMetric] at (-7.05, -8.8) (Pm)
        {\emph{Outcome:} base scorer reproduced, so downstream null results
         are not attributed to pipeline failure.};

  \node[bioLeaf]   at (-2.35, -4.6) (C1)
        {\textbf{Phase-0 noise floor} -- $N{=}200$ size-matched surface
         perturbations per tuple across 6 POIs.};
  \node[bioLeaf]   at (-2.35, -6.7) (C2)
        {\textbf{Sensitivity ablation} -- perturbation count, displacement
         size, and chemistry-aware vs geometric perturbations.};
  \node[bioMetric] at (-2.35, -8.8) (Cm)
        {\emph{Outcome:} per-POI $\sigma$ is \textbf{0.037--0.18};
         the null is POI-specific and chemistry-sensitive.};

  \node[bioLeaf]   at ( 2.35, -4.6) (O1)
        {\textbf{Real \dpt\ operator} -- 15 POIs / 189 interface sites with
         phospho, alanine-scan, and Kme3 perturbations.};
  \node[bioLeaf]   at ( 2.35, -6.7) (O2)
        {\textbf{Cross-checks} -- PROTAC-STAN control, Boltz-2-conditioned
         route on 2 POIs, and dose-response Ala control.};
  \node[bioMetricNeg] at ( 2.35, -8.8) (Om)
        {\emph{Result (negative):} 14.5/15.9/15.7\% clear vs 13.4\% chance;
         \textbf{0/69} survive FDR/Bonferroni.};

  \node[bioLeaf, draw=cBioFroz!60, densely dashed] at ( 7.05, -4.6) (F1)
        {\textbf{Validation track} -- calibrated \dpt\ phospho-PROTAC ranking;
         frozen top-$K$ AUC threshold.};
  \node[bioLeaf, draw=cBioFroz!60, densely dashed] at ( 7.05, -6.7) (F2)
        {\textbf{Ablations and robustness} -- calibration, route comparison,
         scorer comparison, PTM type, and mutant tracks.};
  \node[bioMetric, draw=cBioFroz!50, densely dashed] at ( 7.05, -8.8) (Fm)
        {\emph{Status:} six frozen-threshold experiments await a qualifying
         PTM-sensitive scorer.};

  \foreach \a/\b in {P/P1, P1/P2, P2/Pm,
                     Ca/C1, C1/C2, C2/Cm,
                     Co/O1, O1/O2, O2/Om}{
    \draw[ar] (\a.south) -- (\b.north);
  }
  \foreach \a/\b in {F/F1, F1/F2, F2/Fm}{
    \draw[ar, dashed, cBioFroz!70] (\a.south) -- (\b.north);
  }

  \draw[ar, dashed, gray!60]
        (Cm.south) .. controls +(1.8,-1.3) and +(-1.8,-1.3) .. ([xshift=-4mm]Om.south);
  \node[font=\scriptsize\itshape, color=black!60, align=center] at (0, -10.65)
        {calibrated floor\\defines the null};

  \draw[ar, dashed, cBioFail!60]
        ([xshift=4mm]Om.south) .. controls +(1.8,-1.3) and +(-1.8,-1.3) .. (Fm.south);
  \node[font=\scriptsize\itshape, color=cBioFail!75, align=center] at (4.7, -10.65)
        {negative bound\\defines the benchmark};

\end{tikzpicture}%
}
\caption{Experiment suite for the biomedical drug discovery case.}
\label{fig:bio-exp-taxonomy}
\end{figure}
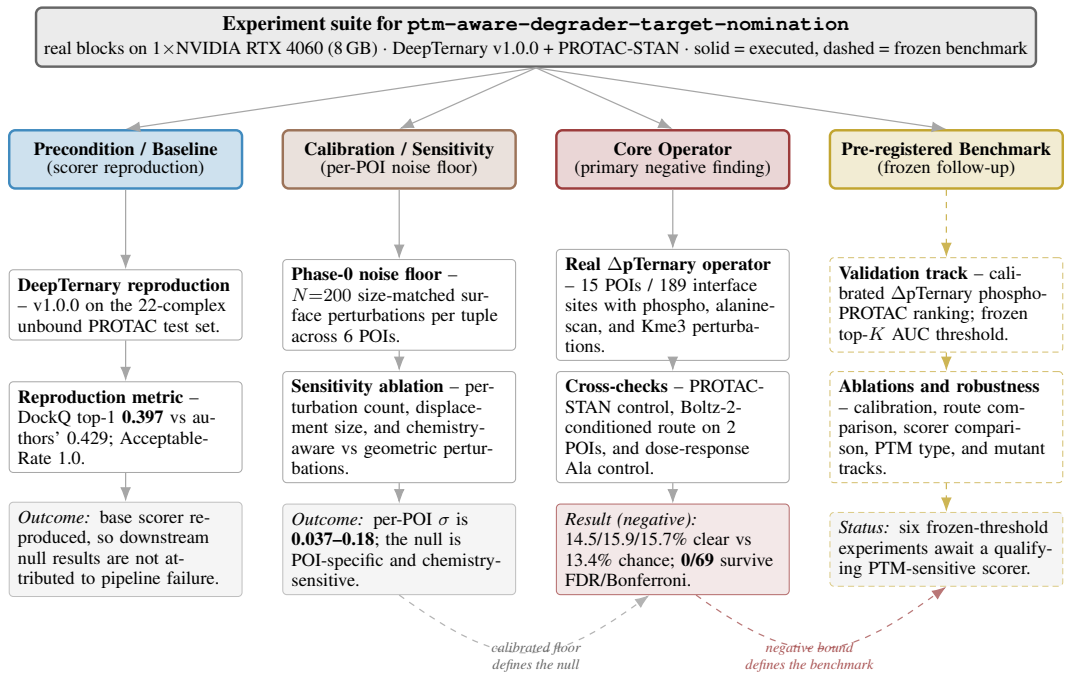
\FloatBarrier

\end{document}